\documentclass[10pt]{article}

\usepackage[a4paper,margin=2.5cm]{geometry}
\usepackage[T1]{fontenc}
\usepackage[utf8]{inputenc}
\usepackage{lmodern}

\usepackage{graphicx}
\usepackage{indentfirst,csquotes}
\usepackage{amssymb,amsthm,amsmath}
\usepackage{xcolor,paralist,hyperref,fancyhdr,etoolbox}
\usepackage{multicol}
\usepackage{multirow}
\usepackage{comment}
\usepackage{algorithm}
\usepackage{algpseudocode}
\usepackage{authblk}

\hypersetup{
  colorlinks=true,
  linkcolor=black,
  filecolor=black,
  urlcolor=black
}

\title{\textbf{Quantifying Gate Contribution in Quantum Feature Maps for Scalable Circuit Optimization}}

\author[1]{F. Rodríguez-Díaz}
\author[2]{D. Gutiérrez-Avilés}
\author[1]{A. Troncoso}
\author[1]{F. Martínez-Álvarez}

\affil[1]{Data Science and Big Data Lab, Pablo de Olavide University, Seville, ES 41013, Spain\\
\texttt{froddia@upo.es, atrolor@upo.es, fmaralv@upo.es}}

\affil[2]{Department of Computer Science, University of Seville, Seville, ES 41012, Spain\\
\texttt{dgutierrez3@us.es}}

\date{\today}

\begin{document}
\maketitle

\begin{abstract}
\noindent Quantum machine learning offers promising advantages for classification tasks, but noise, decoherence, and connectivity constraints in current devices continue to limit the efficient execution of feature map-based circuits. Gate Assessment and Threshold Evaluation (GATE) is presented as a circuit optimization methodology that reduces quantum feature maps using a novel gate significance index. This index quantifies the relevance of each gate by combining fidelity, entanglement, and sensitivity. It is formulated for both simulator/emulator environments, where quantum states are accessible, and for real hardware, where these quantities are estimated from measurement results and auxiliary circuits. The approach iteratively scans a threshold range, eliminates low-contribution gates, generates optimized quantum machine learning models, and ranks them based on accuracy, runtime, and a balanced performance criterion before final testing. The methodology is evaluated on real-world classification datasets using two representative quantum machine learning models, PegasosQSVM and Quantum Neural Network, in three execution scenarios: noise-free simulation, noisy emulation derived from an IBM backend, and real IBM quantum hardware. The structural impact of gate removal in feature maps is examined, compatibility with noise-mitigation techniques is studied, and the scalability of index computation is evaluated using approaches based on density matrices, matrix product states, tensor networks, and real-world devices. The results show consistent reductions in circuit size and runtime and, in many cases, preserved or improved predictive accuracy, with the best trade-offs typically occurring at intermediate thresholds rather than in the baseline circuits or in those compressed more aggressively.
\end{abstract}

\bigskip

\section{Introduction}\label{sec1}

Quantum machine learning (QML) has emerged as a disruptive field that exploits the principles of quantum mechanics to process information in ways unattainable by classical computers \cite{RODRIGUEZ26}. Quantum computers benefit from physical phenomena such as superposition, entanglement, and quantum interference to solve complex problems in cryptography, optimization, or material science exponentially faster than their classical counterparts \cite{Nielsen10}. The advent of Noisy Intermediate-Scale Quantum (NISQ) devices marks a critical step toward realizing the quantum advantage, in which quantum computers outperform classical systems on specific tasks \cite{Preskill18}. 

At the heart of quantum computing lies the quantum circuit model, in which computations are performed by a sequence of quantum gates acting on qubits, the fundamental units of quantum information \cite{Barenco95}. Quantum circuits translate algorithms into physical operations that can be executed on quantum hardware. However, current quantum devices are limited by qubit decoherence, gate errors, and restricted qubit connectivity \cite{Devitt13}. These constraints make it imperative to optimize quantum circuits to minimize errors and resource usage. Optimizing circuits involves reducing the number of gates, shortening the circuit depth, and enhancing fidelity, collectively improving execution time and computational accuracy \cite{Maslov17}. Efficient optimization of quantum circuits is crucial for maximizing the performance of NISQ devices and developing scalable quantum computing \cite{DeJong20}.

Despite significant progress, one of the persistent challenges in quantum computing is mitigating the effects of noise and errors inherent in quantum systems \cite{Kandala19}. Traditional optimization techniques often focus on circuit depth minimization without a holistic approach to improving computational accuracy. Moreover, the stochastic nature of quantum operations requires methodologies that balance resource optimization with the preservation of algorithmic integrity \cite{Murali19}. Optimizing quantum circuits is a critical task in quantum computing, aiming to make quantum algorithms more practical and efficient on current hardware \cite{PERETZ24}. However, several limitations persist that require new methodologies:

\begin{enumerate}
\item Many optimization techniques are designed for specific gate sets or quantum architectures.

\item Some existing techniques focus on reducing circuit depth through gate cancellation and commutation, but do not necessarily assess the individual impact of each gate on computational accuracy.

\item Hardware-aware methods optimize qubit allocation and routing to suit specific quantum architectures. While they improve execution efficiency, they do not evaluate gate significance (or gate individual contribution and meaningfulness to the circuit) within the circuit's computational context.

\item Error mitigation strategies, including noise-adaptive compilation and variational error reduction, adjust circuits to minimize errors, but often increase circuit complexity or require additional quantum resources, such as ancillary qubits \cite{ANDERS10} or extended gate sequences \cite{XU13}.

\item Existing methods do not provide a unified metric to quantify the significance of individual gates for resource optimization and computational accuracy. This gap limits the ability to decide which gates can be safely removed without degrading performance.
\end{enumerate}

This research is motivated by the need to develop the Gate Assessment and Threshold Evaluation (GATE) methodology to optimize quantum circuits by improving their accuracy and execution efficiency. GATE relies on a novel Gate Significance Index (GSI), a novel metric that evaluates the impact of each gate in the quantum circuit on the overall computational outcome, and a systematic method for determining optimal thresholds to remove gates with minimal contribution to circuit performance. Thus, our contribution to optimizing QML circuit performance and execution time through the removal of insignificant gates can be summarized through several key advantages:

\begin{enumerate}

\item A novel methodology to optimize quantum circuits by reducing the number of gates, thus improving the accuracy and reducing the execution time.

\item Unlike methods that focus solely on circuit depth, GSI enables the assessment of each gate's importance within the computation. This enables selective gate elimination based on quantitative significance.

\item When we focus on gates that contribute little to the desired results, we reduce the circuit's source requirements and often improve computational accuracy. Eliminating insignificant gates decreases the cumulative error introduced by noisy operations, a benefit not explicitly addressed in prior works.

\item Our approach can be integrated with other optimization strategies, such as hardware-aware mapping and error mitigation techniques. When you incorporate this additional layer of optimization, the overall circuit performance improves beyond what existing methods alone can achieve.
\end{enumerate}

To validate the effectiveness of our optimization methodology, we conducted extensive experiments on nine datasets using both the PegasosQSVM algorithm, a Quantum Support Vector Machine (QSVM) \cite{SIMON24} that employs the Pegasus architecture as a classifier \cite{SCHULD17}, and a Quantum Neural Network (QNN) architecture \cite{Hirai24}. The optimized circuits showed a substantial reduction in gate count (up to 40\% in some cases) and achieved faster execution times. Additionally, the algorithm's accuracy was preserved and even improved on most datasets.

In addition, to validate the adaptability of the GSI methodology, we extended our experiments to include simulations using a realistic noise backend that models the behavior of real quantum devices. We run this simulator on a GPU to accelerate temporal execution across the different models. The results demonstrate that GSI can efficiently adjust to the topological constraints and noise characteristics of real hardware. Furthermore, a noise-mitigation technique based on dynamic decoupling \cite{Ramadhani23} was incorporated to show the independence of the GATE methodology when running on actual hardware. 

The rest of the paper is structured as follows. Section \ref{sec:relatedWork} provides a detailed review of related work in the field of quantum circuit optimization. Section \ref{sec:methodology} introduces the methodology, including the process of calculating GSI and the systematic removal of gates based on it. Section \ref{sec:results} presents the experimental results, demonstrating the effectiveness of our optimization approach in various datasets. Section \ref{sec:scalabilityResults} reports the results obtained from scalability tests for calculating GSI values using different methods. Section \ref{sec:conclusion} summarizes our findings, discusses the limitations of the current approach, and suggests potential future works. Finally, we provide supplementary material with full access to the code and datasets used in this study.

\section{Related Work}\label{sec:relatedWork}
In this section, we review works related to quantum circuit optimization considering several key features: reduction of circuit depth (F1), error mitigation strategies (F2), hardware-aware optimization techniques (F3), evaluation of individual gate contributions (F4), scalability to larger systems (F5), applicability on various algorithms (F6), balance between resource optimization and accuracy (F7), and the use of a quantitative gate significance metric (F8). 

One of the foundational works in quantum circuit design is the paper by Barenco et al. \cite{Barenco95}, which introduced a set of elementary gates for quantum computation, laying the groundwork for constructing complex quantum operations from simpler components. Nielsen and Chuang expanded on these principles, providing essential knowledge on quantum computation and information \cite{Nielsen10}.

Maslov and Miller introduced techniques for reducing the depth in reversible and quantum circuits \cite{Maslov08}. They significantly reduced circuit complexity by applying transformation rules and exploiting gate commutation. Their work provides foundational methods for simplifying quantum circuits without altering their functionality.

Nam et al. proposed automated optimization passes for quantum transpilers that systematically apply gate cancellation and commutation rules \cite{Nam18}. These transpiler optimizations reduce gate counts and circuit depth by identifying and eliminating redundant operations, which is particularly effective for circuits implementing quantum algorithms like the Quantum Fourier Transform \cite{GRIFFITHS96}. This improves execution efficiency on quantum hardware by reducing the total number of operations.

Optimizing quantum circuits often involves reducing the depth to minimize error accumulation and resource usage. Amy et al. developed an algorithm for synthesizing depth-optimal quantum circuits, with a primary focus on reducing the T-gate count in Clifford+T circuits \cite{Amy13}. The method efficiently finds optimal circuits by combining forward and backward search techniques, significantly reducing the computational resources compared to exhaustive search methods. This approach is crucial for fault-tolerant quantum computing, where minimizing non-Clifford gates like T-gates reduces the error correction overhead. Maslov \cite{Maslov17} discussed circuit compilation techniques specifically for ion-trap quantum machines, highlighting methods to minimize circuit depth. Shende et al. explored the synthesis of quantum logic circuits and provided algorithms for efficient circuit construction \cite{Shende06}.

Error mitigation is crucial in the NISQ era. Hence, Temme et al. proposed methods to mitigate errors in short-depth quantum circuits via zero-noise extrapolation \cite{Temme17}. Kandala et al. demonstrated that error mitigation techniques could extend the computational reach of noisy quantum processors, emphasizing the importance of such strategies in practical quantum computing \cite{Kandala19}. Li and Benjamin \cite{Li17} highlighted the role of circuit optimization in error mitigation and reduced the exposure of quantum computations to noise, enhancing the reliability of results. 

Efficiently mapping quantum circuits into hardware with limited connectivity and high error rates is a significant challenge. Murali et al. \cite{Murali19} presented noise-adaptive compiler mappings for NISQ devices, adapting circuit compilation to hardware-specific noise characteristics. Siraichi et al. addressed qubit allocation by considering the connectivity constraints of quantum hardware \cite{Siraichi18}. The proposed methods optimize qubit placement and quantum-operation routing to minimize the need for SWAP gates, which are costly in both time and error rate. By reducing the number of SWAP gates, they decreased circuit depth and execution time, improving performance on hardware with limited qubit connectivity.

Variational quantum algorithms, which adjust parameters to minimize a cost function, have also been used to optimize circuits. Mitarai et al. \cite{Mitarai18} proposed a method for constructing optimized variational quantum circuits for machine learning tasks. Bravy et al. \cite{Bravy22} discussed hybrid quantum-classical algorithms for parameter optimization in quantum circuits. Khatri et al. \cite{Khatri19} presented an algorithm, quantum-assisted quantum compilation, based on a variational approach in which the circuit structure and parameters are optimized. They designed circuits to perform optimally on given quantum devices by incorporating hardware-specific constraints and noise models. 

Coles et al. \cite{Coles18} provided implementations of quantum algorithms for beginners, offering useful information on practical aspects of quantum computing and optimization. Similarly, Schuld et al. explored the implementation of quantum machine learning algorithms, highlighting the need for optimized circuits in computational tasks \cite{Schuld15}.

In the context of mapping and compiling qubits, ADAC \cite{Huang25} addresses the NP-complete problem of qubit mapping by combining structure-aware circuit partitioning (using subgraph isomorphism checks) with a heuristic routing strategy, which significantly reduces SWAP and maintains practical overhead for small and medium-sized circuits. Another study \cite{Rattacaso25} proposes performing compilation using quantum computers through procedures based on quantum and simulated cooling, demonstrating scalability in Trotterized Hamiltonian simulation (up to 64 qubits and 64 time steps) and Quantum Fourier Transform (up to 40 qubits and 771 time steps) with fidelity gains that increase with circuit size in translation-invariant environments. Finally, the authors in \cite{Sun25} accelerate optimal mapping based on solvers by applying machine learning-guided reduction at both the global and local levels and increasing training data through gate assignment and qubit reorganization, resulting in an average acceleration of 1.79 times and an advantage of approximately 22\% in spatial complexity compared to the most advanced approaches.

In Valois et al. \cite{Valois26}, qubit assignment is solved exactly with a branch-and-bound method formulated as a quadratic assignment problem, and scalability is extended through HPC parallelization, enabling optimal solutions for reference circuits up to 26 qubits. And in the work \cite{Yale25}, trapped ion compilation is adapted to the gate set and noise by using continuously parameterized ZZ gates, utilizing optimizations such as mirror swapping, pair selection for best-performing links, and circuit approximation to eliminate low-impact ZZ gates, improving observed performance (e.g., quantum volume) and analyzing the effects of stochastic versus coherent errors.

Table \ref{tab:Related} summarizes the contributions of the works reviewed according to the eight features considered at the beginning of this section.

\begin{table}[htb]
\centering
\caption{Summary of the revised works.}
\label{tab:Related}
\setlength{\tabcolsep}{4pt}
\renewcommand{\arraystretch}{1.2}
\begin{tabular}{|l|c|c|c|c|c|c|c|c|}
\hline
\textbf{Ref.} & \textbf{F1} & \textbf{F2} & \textbf{F3} & \textbf{F4} & \textbf{F5} & \textbf{F6} & \textbf{F7} & \textbf{F8} \\
\hline
\cite{Barenco95}  & \checkmark &            &            &            &            &            &            &            \\
\cite{Nielsen10}  &            &            &            &            &            &            &            &            \\
\cite{Maslov08}   & \checkmark &            &            &            &            &            &            &            \\
\cite{Nam18}      & \checkmark &            &            &            & \checkmark & \checkmark &            &            \\
\cite{Amy13}      &            &            &            &            &            &            &            &            \\
\cite{Maslov17}   & \checkmark &            & \checkmark &            &            &            &            &            \\
\cite{Shende06}   &            &            &            &            &            &            &            &            \\
\cite{Temme17}    &            & \checkmark &            &            &            &            &            &            \\
\cite{Kandala19}  &            & \checkmark &            &            &            &            &            &            \\
\cite{Li17}       & \checkmark & \checkmark &            &            &            &            &            &            \\
\cite{Murali19}   &            & \checkmark & \checkmark &            &            &            &            &            \\
\cite{Siraichi18} & \checkmark &            & \checkmark &            &            &            &            &            \\
\cite{Mitarai18}  &            &            &            &            & \checkmark & \checkmark &            &            \\
\cite{Bravy22}    &            &            &            &            & \checkmark & \checkmark &            &            \\
\cite{Khatri19}   &            & \checkmark &            &            &            &            &            &            \\
\cite{Coles18}    &            &            &            &            &            & \checkmark &            &            \\
\cite{Schuld15}   &            &            &            &            &            & \checkmark &            &            \\

\cite{Valois26} &            &            &            &            & \checkmark & \checkmark &  &            \\
\cite{Yale25}  &            & \checkmark & \checkmark & \checkmark &            &            &            &            \\
\cite{Sun25}      &            &            & \checkmark &            &            & \checkmark &  &            \\
\cite{Rattacaso25}  &            &            &            &            & \checkmark & \checkmark &  &            \\
\cite{Huang25}      & \checkmark &            & \checkmark &            & \checkmark &            &            &            \\
\hline

\textbf{GSI}      & \checkmark & \checkmark & \checkmark & \checkmark & \checkmark & \checkmark & \checkmark & \checkmark \\
\hline
\end{tabular}
\end{table}

Given the previously discussed works, the need for efficient quantum circuit optimization remains critical as the field advances toward practical quantum computing. Our proposed methodology fills a gap in existing research by providing a quantitative means of evaluating and optimizing quantum circuits at the gate level. By introducing the GSI, we enable the selective removal of insignificant gates, reducing resource requirements and enhancing computational accuracy without the drawbacks associated with other approaches. This work differentiates itself by focusing on the individual contribution of gates, balancing optimization with accuracy, and offering a scalable, hardware-agnostic solution that complements existing optimization techniques. The computation of the proposed GSI introduces a novel strategy for evaluating individual gate contributions in a quantum circuit. Given the nature of existing work and the limited availability of code and data, to the author's knowledge, it is not possible to directly compare our proposal with others. 

\section{Methodology}\label{sec:methodology}

This section introduces the proposed methodology to optimize quantum circuits. Section \ref{sec:GSI} first presents the GSI, a new index designed to assess the individual significance of each gate within a quantum circuit. The methodology, which incorporates GSI as a critical step for optimizing quantum circuits, is introduced in Section \ref{sec:GATE}.

\subsection{Defining the Gate Significance Index}\label{sec:GSI}
In this section, we introduce the Gate Significance Index (GSI), a metric that quantifies the contribution of each gate within a quantum circuit. The GSI is calculated for each gate based on three key metrics: fidelity, entanglement, and sensitivity. The rationale for combining these metrics lies in their ability to collectively capture the essential aspects of a gate’s role in a quantum circuit by ensuring computational accuracy (fidelity), contributing to quantum resources like entanglement (entanglement), and measuring the gate’s overall impact on the system (sensitivity). Together, these metrics are likely to provide a comprehensive and balanced framework for evaluating gate significance, addressing key trade-offs between functionality, resource utilization, and noise mitigation. 

The metrics comprising the GSI are analyzed below. On the one hand, their formulation in simulation and emulation is presented, allowing access to the quantum state and direct calculation of quantities such as fidelity, partial trace, and sensitivity. On the other hand, the formulation for real hardware is introduced, in which the state is inaccessible and metrics must be estimated from measurement results (counts) and auxiliary circuits. This distinction allows the same interpretation of the GSI to be maintained, while operational definitions are adapted to experimental limitations and the effects of noise.

\subsubsection{GSI formulation in simulator/emulator}

Below, we detail how each component of the GSI is computed in simulation and emulation, in which the quantum state can be accessed (explicitly or implicitly) during circuit evolution, enabling direct evaluation of state-based quantities. We begin with the fidelity term, which captures how strongly the quantum state changes after applying each gate.

\begin{enumerate}
\item  Fidelity (F) \cite{VADALI24} measures the similarity between two quantum states. In the context of a quantum gate, it quantifies the extent to which the state changes after the gate is applied. A fidelity close to 1 indicates that the gate preserves the state, whereas a low fidelity suggests that the gate significantly transforms it. This enables the identification of key gates that are essential for significantly transforming the quantum state. The fidelity for a couple states \(\rho\) and \(\sigma\) is calculated as Eq. (\ref{eq:F}):

    \begin{equation} \label{eq:F}
        F(\rho, \sigma) = \text{Tr} \left( \sqrt{ \sqrt{\rho} \sigma \sqrt{\rho} } \right)
    \end{equation}

\noindent where \(\rho\) and \(\sigma\) are the density matrices of the quantum states being compared and \text{Tr} represents the trace operation, which sums the diagonal elements of a matrix, providing a scalar that helps quantify the similarity between \(\rho\) and \(\sigma\).

\item  Entanglement (E) \cite{KOOKANI24} is a quantum property in which the states of two or more qubits become interconnected, such that the state of one qubit cannot be described independently of the state of the others. In the context of quantum gates, entanglement enables the creation of quantum correlations essential for many quantum algorithms, as it allows information to be processed and shared across the qubits in ways that classical systems cannot replicate. A gate contributes to the quantum advantage by generating or enhancing entanglement and enabling more complex, efficient computations that leverage these correlations. This metric helps identify gates that are pivotal for entanglement generation, enabling the design of more efficient circuits while preserving their computational power. 

For entanglement analysis, the full quantum system is partitioned into two complementary subsystems, denoted \(A\) and \(B\). Subsystem \(A\) may represent a single qubit or a subset of qubits, while subsystem \(B\) contains the remaining qubits. Given the pure state \(|\psi\rangle\) of the full system, the reduced density matrix of subsystem \(A\) is obtained by tracing out subsystem \(B\), that is, \(\rho_A = \mathrm{Tr}_B(|\psi\rangle\langle\psi|)\). The entanglement between \(A\) and \(B\) is then quantified through the von Neumann entropy of \(\rho_A\) (Eq. \ref{eq:entaglement}):

 \begin{equation}\label{eq:entaglement}
        E(|\psi\rangle) = S(\rho_A) = -Tr(\rho_A \log_2 \rho_A)
\end{equation}

where \(\mathrm{Tr}(\cdot)\) denotes the trace operation and \(\log_2\) is the base-2 logarithm, which means that the entropy is measured in bits. For a bipartite pure state, this quantity reflects the quantum correlations between subsystems \(A\) and \(B\): a value of zero indicates that the state is separable, whereas larger values indicate stronger entanglement and a greater degree of non-separability between the two subsystems.

\item  Sensitivity (P) \cite{BU24} evaluates how the circuit responds to small variations in gate parameters or inputs. A highly sensitive gate can amplify small errors, which impacts the circuit's robustness and stability. Identifying and adjusting highly sensitive gates enhances fault tolerance, enabling the design of circuits that are less susceptible to noise and physical variations. Sensitivity is defined as Eq. (\ref{eq:sensitivity}):

    \begin{equation}\label{eq:sensitivity}
        P = \left| \frac{\partial \langle O \rangle}{\partial \theta} \right|
    \end{equation}

where $\langle O \rangle$ denotes the expected value of an observable $O$ for the quantum state, and $\theta$ is the parameter associated with a parameterized gate. Sensitivity is calculated only for gates that contain adjustable parameters. In such cases, the method creates a copy of the current circuit and perturbs the parameter of the last parameterized gate according to the values assigned to the gate’s corresponding weights. For each perturbed version of the circuit, the fidelity with respect to the original state is evaluated, providing an indirect measure of how variations in the gate parameter affect the final quantum state. The standard deviation of these fidelity values is then used as a measure of sensitivity. This does not mean that fidelity directly represents the observable $\langle O \rangle$; rather, it acts as an indicator to quantify the circuit’s stability in the face of parameter perturbations. A sensitivity value close to 1 indicates that small changes in $\theta$ produce large variations in the final state, which can lead to instability or error amplification. Conversely, a value close to 0 indicates that variations in the parameters have little effect, resulting in a more stable and robust gate. For non-parameterized doors, such as gate $H$, sensitivity is not calculated and a default value of 0 is assigned. Therefore, only fidelity and entanglement are evaluated, whereas parameterized gates are evaluated using fidelity, entanglement, and sensitivity.

\end{enumerate}

\noindent These three metrics form the basis for the GSI, which quantifies the overall significance of each individual gate. The GSI is computed by averaging such metrics, as shown in Eq. (\ref{eq:GSI}):

    \begin{equation}\label{eq:GSI}
    \begin{split}
         GSI = \frac{F + E + (1 - P)}{3}  \\
         %= \frac{2 + \text{Tr} \left( \sqrt{ \sqrt{\rho} \sigma \sqrt{\rho} } \right) - Tr(\rho_A \log_2 \rho_A) + Tr(\rho \log_2 \rho) - \left| \frac{\partial \langle O \rangle}{\partial \theta} \right|}{4}            
    \end{split}
    \end{equation}

where $GSI \in [0,1]$, since all three metrics considered also range from 0 to 1. 
On the one hand, F and E are metrics that quantify positive contributions to a gate's significance.  GSI prioritizes gates that maintain computational correctness and make use of valuable quantum resources, such as entanglement, by rewarding their importance in the circuit with the sum $F+E$. However, P is a factor that can reduce the effectiveness of a gate, making it less effective. The GSI penalizes gates that introduce excessive noise or instability using $(1-P)$. This approach rewards gates that maintain stability, ensuring that they contribute positively to the circuit. So, the formula balances positive contributions $(F+E)$ with the absence of negative effects $(1-P)$, ensuring that the GSI represents a comprehensive evaluation of the relevance of a gate. Dividing by 3 averages these components to normalize the index, making it consistent across different circuits and applications.

The pseudocode for the calculation of GSI values is introduced in Pseudocode \ref{alg:calc_gsi}.

\begin{algorithm}
\caption{Calculate GSI}
\label{alg:calc_gsi}
\scriptsize
\begin{algorithmic}[1]
\Require \textit{feature\_map}, \textit{parameter\_binds}, \textit{method} $\in\{\texttt{density\_matrix},\texttt{matrix\_product\_state}, \texttt{tensorial\_network}\}$, \textit{ent\_qubit}, $deltas=\{0,+\delta,-\delta\}$
\Ensure \textit{metrics}

\State $qc \gets \textsc{Decompose}(\textsc{AssignParameters}(\textit{feature\_map},\ \textit{parameter\_binds}))$
\State $n \gets qc.num\_qubits$
\State $\textit{ent\_qubit} \gets \min(\textit{ent\_qubit}, n-1)$
\State $sim \gets \textsc{AerSimulator}(\textit{method})$
\State $tmp \gets \textsc{QuantumCircuit}(n)$
\State $gates \gets \emptyset$

\ForAll{$(i,\ inst)\in qc.data$}
  \State $gate \gets inst.operation$
  \State $qidx \gets \textsc{IndicesOf}(inst.qubits)$
  \State \textsc{Append}$(gates,\ (gate,qidx))$

  \State $\textsc{SaveDensityMatrix}(tmp,\ qubits=qidx,\ label=\texttt{"rho\_prev\_"}i)$
  \State $\textsc{Append}(tmp,\ gate,\ qidx)$
  \State $\textsc{SaveDensityMatrix}(tmp,\ qubits=\{\textit{ent\_qubit}\},\ label=\texttt{"rho\_ent\_"}i)$
\EndFor

\State $data \gets \textsc{Run}(sim,\ tmp)$
\State $metrics \gets \emptyset$

\ForAll{$(i,\ (gate,qidx)) \in gates$}
  \State $\rho_{\mathrm{prev}} \gets data[\texttt{"rho\_prev\_"}i]$
  \State $U \gets \textsc{Operator}(gate)$
  \State $F \gets \left|\mathrm{Tr}\!\left(\rho_{\mathrm{prev}}\, U\right)\right|^2$

  \State $\rho_{\mathrm{ent}} \gets data[\texttt{"rho\_ent\_"}i]$
  \State $E \gets \textsc{Entropy}(\rho_{\mathrm{ent}})$
  \State $E_{\mathrm{norm}} \gets E / \log_2(\dim(\rho_{\mathrm{ent}}))$

  \If{$gate$ is parameterized}
     \State $P \gets \Call{CalculateSensitivityOnGateSupport}{\rho_{\mathrm{prev}},\ gate,\ deltas}$ \Comment{Alg.~\ref{alg:calc_sensitivity}}
  \Else
     \State $P \gets 0$
  \EndIf

  \State $M \gets \dfrac{F + E_{\mathrm{norm}} + (1 - P)}{3}$
  \State \textsc{Append}$(metrics,\{\text{'gate'}{:}gate.name,\ \text{'position'}{:}i,\ \text{'F'}{:}F,\ \text{'E'}{:}E_{\mathrm{norm}},\ \text{'S'}{:}S_{\mathrm{norm}},\ \text{'P'}{:}P,\ \text{'M'}{:}M\})$
\EndFor

\Return $metrics$
\end{algorithmic}
\end{algorithm}

First, the code assigns values to the circuit parameters, decomposes it into elementary gates, and prepares a single run using the calculation method selected and explained in Section \ref{sec:foundationsScalability}. Instead of extracting intermediate quantum states through explicit step-by-step simulation of the circuit, the simulator is designed to store only the reduced-density matrices necessary for evaluating the GSI components. Specifically, before applying each gate at position $i$, the algorithm stores the reduced-density matrix in the qubits on which that gate acts, which is then used to calculate both fidelity and sensitivity at the local level. After applying the gate, it stores the reduced-density matrix of a fixed qubit.

\begin{algorithm}
\caption{Calculate Sensitivity}
\label{alg:calc_sensitivity}
\begin{algorithmic}[1]
\Require $\rho_{\mathrm{prev}}$ (reduced density matrix on the gate support before applying the gate), $gate$ (parameterized), $deltas=\{0,+\delta,-\delta\}$
\Ensure $P$

\State $\theta \gets \textsc{FirstParam}(gate)$
\State $U_{\mathrm{base}} \gets \textsc{Operator}(gate)$
\State $fids \gets \varnothing$

\For{$\Delta \in deltas$}
    \State $gate' \gets \textsc{Copy}(gate)$
    \State $\textsc{SetFirstParam}(gate',\ \theta + \Delta)$
    \State $U_{\mathrm{pert}} \gets \textsc{Operator}(gate')$
    \State $V \gets U_{\mathrm{base}}^\dagger\, U_{\mathrm{pert}}$
    \State $f \gets \left|\mathrm{Tr}\!\left(\rho_{\mathrm{prev}}\, V\right)\right|^2$
    \State \textsc{Append}$(fids,\ f)$
\EndFor

\State $P \gets \textsc{Std}(fids)$
\State \Return $P$

\end{algorithmic}
\end{algorithm}

Once the simulator is complete, all quantities are calculated by post-processing these saved reduced states. The fidelity in step $i$ is obtained from the reduced state before the gate 
$\rho_{prev,i}$ and the gate unit $U_i$ through the local superposition $F_i = |Tr(\rho_{prev, i}U_i)|^2$. The entanglement term is calculated as the normalized von Neumann entropy of the reduced state of a qubit saved after applying the gate. Sensitivity is calculated by evaluating a small set of local superpositions for parameter perturbations $\theta \rightarrow \theta + \delta$ using the same pre-gate reduced state $\rho_{prev,i}$ and the local operator $V(\delta) = U(\theta)^\dagger U(\theta, \delta)$. The sensitivity $P_i$ is then the standard deviation over these superposition values. Finally, the algorithm combines the terms in the GSI score and stores the metrics for each gate.

In terms of computational complexity, let $n$ be the number of qubits and $N$ the number of elementary gates after decomposition. The dominant cost is the single execution of AerSimulator on a circuit of depth $N$ using the chosen method. In Density Matrix simulation (DM), the state scales as $4n$, so memory and time grow exponentially with $n$. In Matrix Product State (MPS) simulation, the cost is determined by the maximum link dimension $x$, with memory typically scaling as $O(nx2)$ and a two-qubit gate application costing approximately $O(x^3)$, resulting in a total execution time that is often written as $O(Nx^3)$ when $x$ remains bounded, but which can degrade towards exponential behavior in high-entanglement regimes where $x$ grows rapidly. In Tensor Network (TN) simulation, the execution time is dominated by the contraction of a network whose cost depends on the contraction path and the induced tree width $w$, with a worst-case scale of the order of $O(poly(N)2^w)$ for both time and memory, where $w$ can approach $n$ for dense/highly intertwined connectivity and thus effectively become exponential. In contrast, more structured circuits (e.g., close to 1D or weakly intertwined) can admit a small $w$ and thus much better scaling. The post-processing stage is linear in the number of gates, $O(N)$, because each gate contributes a constant amount of work using reduced states defined on the gate support (typically 1-2 qubits) and on a fixed reduced state of one qubit for entanglement, and for sensitive parameterized gates.

\subsubsection{GSI formulation in real hardware}\label{sec:GSIRH}

Next, we show how each component of the GSI is estimated in actual quantum hardware. Unlike simulation/emulation, the quantum state cannot be accessed directly, and all quantities must be inferred from measurement outcomes. To do this, we replace state-based expressions with operational estimators constructed from shot statistics and auxiliary circuits.

\begin{enumerate}
\item Fidelity estimator ($\widehat{F}$).
For each gate position $i$, let $U_i$ denote the prefix circuit up to (and including) the $i$-th gate, and $U_{i-1}$ the previous prefix. We construct the overlap circuit in Eq. (\ref{eq:overlap_circuit_hw}).

\begin{equation}
\label{eq:overlap_circuit_hw}
V_i = U_{i-1}^{\dagger}U_i,
\end{equation}

measure it in the computational basis, and estimate the fidelity contribution as the probability of obtaining the all-zero outcome expressed in Eq. (\ref{eq:fidelity_hat_hw}).

\begin{equation}
\label{eq:fidelity_hat_hw}
\widehat{F}_i = p(0\cdots 0 \mid V_i).
\end{equation}

This estimator corresponds to $|\langle \psi_{i-1}|\psi_i\rangle|^2$ in the noiseless limit, while in hardware it naturally incorporates the effect of device noise and finite-shot sampling.

\item Entanglement estimator ($\widehat{E}$).
Since obtaining a reduced density matrix via partial trace is not feasible on hardware, we estimate a \emph{single-qubit} reduced state for a fixed qubit $q$ (e.g., $q=1$) using Pauli tomography. For each prefix $U_i$, we run three measurement settings to estimate the Bloch components as shown in Eq. (\ref{eq:pauli_exp_hw})

\begin{equation}
\label{eq:pauli_exp_hw}
x_i = \langle X_q \rangle,\quad y_i = \langle Y_q \rangle,\quad z_i = \langle Z_q \rangle,
\end{equation}

and reconstruct the one-qubit density matrix Eq. (\ref{eq:rho1q_hw}).

\begin{equation}
\label{eq:rho1q_hw}
\widehat{\rho}_{q,i} = \frac{1}{2}\left(I + x_i X + y_i Y + z_i Z\right).
\end{equation}

The entanglement contribution is then approximated by the von Neumann entropy of this reconstructed reduced state, computed according Eq. (\ref{eq:ent_hat_hw}).

\begin{equation}
\label{eq:ent_hat_hw}
\widehat{E}_i = S(\widehat{\rho}_{q,i}) = -\mathrm{Tr}\left(\widehat{\rho}_{q,i}\log_2 \widehat{\rho}_{q,i}\right).
\end{equation}

Please note that this is an operational proxy, i.e., it captures the degree of mixing achieved by the selected qubit due to correlations with the rest of the register and noise from the machinery, without requiring access to the complete state.

\item Sensitivity estimator ($\widehat{P}$).
For parametrized gates, we avoid re-simulating statevectors and instead use the same overlap construction to quantify the effect of small parameter perturbations. Let $\theta$ be the parameter of the current gate at position $i$, and define two perturbed prefixes $U_i^{+}$ and $U_i^{-}$ where only the last gate parameter is shifted by $\pm\delta$. We compute the corresponding overlap probabilities $\widehat{F}_i^{+}$ and $\widehat{F}_i^{-}$ (via the same all-zero measurement on $U_i^\dagger U_i^{\pm}$) and define as given in Eq. (\ref{eq:sens_hat_hw}).

\begin{equation}
\label{eq:sens_hat_hw}
\widehat{P}_i = \mathrm{std}\left(1,\ \widehat{F}_i^{+},\ \widehat{F}_i^{-}\right).
\end{equation}
For non-parametrized gates, we set $\widehat{P}_i = 0$.

\end{enumerate}

\noindent These estimators yield a hardware-compatible GSI defined in Eq. (\ref{eq:gsi_hardware_hw}).
\begin{equation}
\label{eq:gsi_hardware_hw}
\widehat{GSI}_i = \frac{\widehat{F}_i + \widehat{E}_i + (1 - \widehat{P}_i)}{3}.
\end{equation}

This formulation preserves the interpretation used in simulation/emulation while replacing state-access operations (partial trace, exact fidelity) with measurable quantities derived from counts and auxiliary circuits.

Pseudocode \ref{alg:gsi_hw} summarizes the procedure. The algorithm builds prefix circuits, generates (i) overlap circuits for $\widehat{F}$, (ii) single-qubit Pauli measurement circuits for $\widehat{E}$, and (iii) overlap circuits for the $\pm\delta$ perturbations used in $\widehat{P}$. All circuits are then transpiled to the backend ISA and executed in batches, and the final per-gate metrics are assembled from the observed counts.

\begin{algorithm}
\caption{Calculate hardware GSI (feature map, parameter binds, backend)}
\label{alg:gsi_hw}
\scriptsize
\begin{algorithmic}[1]
\Require \textit{feature\_map}, \textit{parameter\_binds}, \textit{backend}, shots, qubit $q$, $\delta$
\Ensure \textit{metrics}

\State $qc \gets \textsc{AssignParameters}(\textit{feature\_map}, \textit{parameter\_binds})$
\State $qc \gets \textsc{Decompose}(qc)$
\State $U \gets \textsc{Prefixes}(qc)$ \Comment{$U_i$ prefix circuits}
\State $C \gets [\ ]$, $tags \gets [\ ]$

\For{$i = 1$ to $|qc.data|$}
  \State $U_{i-1} \gets \emptyset$ if $i=1$ else $U_{i-1}$
  \State $U_i \gets U_i$

  \State $V_i \gets U_{i-1}^{\dagger} U_i$; \ \textsc{MeasureAll}$(V_i)$
  \State \textsc{Append}$(C, V_i)$; \textsc{Append}$(tags, (\text{"F"}, i))$

  \If{$qc.num\_qubits > 1$}
    \State $(M_Z, M_X, M_Y) \gets \textsc{MeasureXYZSingleQubit}(U_i, q)$
    \State \textsc{Append}$(C, M_Z, M_X, M_Y)$
    \State \textsc{Append}$(tags, (\text{"E\_Z"}, i), (\text{"E\_X"}, i), (\text{"E\_Y"}, i))$
  \EndIf

  \If{$\textsc{HasParameter}(qc.data[i])$}
    \State $U_i^{+} \gets \textsc{ReplaceLastParam}(U_i, \theta+\delta)$
    \State $U_i^{-} \gets \textsc{ReplaceLastParam}(U_i, \theta-\delta)$
    \State $V_i^{+} \gets U_i^{\dagger} U_i^{+}$; \ \textsc{MeasureAll}$(V_i^{+})$
    \State $V_i^{-} \gets U_i^{\dagger} U_i^{-}$; \ \textsc{MeasureAll}$(V_i^{-})$
    \State \textsc{Append}$(C, V_i^{+}, V_i^{-})$
    \State \textsc{Append}$(tags, (\text{"P\_plus"}, i), (\text{"P\_minus"}, i))$
  \EndIf
\EndFor

\State $C_{\text{ISA}} \gets \textsc{TranspileToBackend}(C, backend)$
\State $results \gets \textsc{RunSampler}(C_{\text{ISA}}, shots)$

\ForAll{$(kind, i)$ in $tags$}
  \State \textsc{ParseCounts} and store in maps $\widehat{F}_i$, $(x_i,y_i,z_i)$, $\widehat{F}_i^{+}$, $\widehat{F}_i^{-}$
\EndFor

\For{$i = 1$ to $|qc.data|$}
  \State $\widehat{F}_i \gets p(0\cdots 0 \mid V_i)$
  \If{$qc.num\_qubits > 1$}
    \State $\widehat{\rho}_{q,i} \gets \frac{1}{2}(I + x_i X + y_i Y + z_i Z)$
    \State $\widehat{E}_i \gets S(\widehat{\rho}_{q,i})$
  \Else
    \State $\widehat{E}_i \gets 0$
  \EndIf
  \If{gate $i$ parametrized}
    \State $\widehat{P}_i \gets \mathrm{std}(1,\widehat{F}_i^{+},\widehat{F}_i^{-})$
  \Else
    \State $\widehat{P}_i \gets 0$
  \EndIf
  \State $\widehat{GSI}_i \gets (\widehat{F}_i + \widehat{E}_i + (1-\widehat{P}_i))/3$
  \State \textsc{Append}$(metrics,\{gate, position, \widehat{F}_i,\widehat{E}_i,\widehat{P}_i,\widehat{GSI}_i\})$
\EndFor

\Return \textit{metrics}
\end{algorithmic}
\end{algorithm}

In terms of computational complexity, let $n$ be the number of qubits and $N$ the number of elementary gates after decomposition. The hardware procedure is linear in $N$ in the sense that it generates only a constant number of auxiliary circuits per gate. Concretely, for each gate position $i$ it builds one overlap circuit $V_i=U_{i-1}^\dagger U_i$ to estimate $\widehat{F}_i$ and (when $n > 1$) three additional circuits to estimate $\langle X_q \rangle$, $\langle Y_q \rangle$ and $\langle Z_q \rangle$ for the fixed qubit $q$, which are then used to reconstruct $\widehat{\rho}_{q,i}$ and compute $\widehat{E}_i$. For parametrized gates, it adds two additional overlap circuits to obtain $\widehat{F}_i^{+}$ and $\widehat{F}_i^{-}$ and to compute $\widehat{P}_i$. For non-parametrized gates, these two circuits are not required.

Therefore, the total number of executed circuits scales as $O(N)$. If each circuit is run with a fixed number of shots, the sampling cost scales as $O(shots \times N)$ in terms of total circuit evaluations. The classical post-processing is also $O(N)$, since each gate contributes a constant amount of work. In practice, the observed wall-clock time is dominated not only by this linear scaling but also by backend-dependent overheads such as transpilation/scheduling, queuing latency, and per-circuit execution time on the QPU.

\subsection{Description of the Gate Assessment and Threshold Evaluation methodology}\label{sec:GATE}
This section introduces the novel Gate Assessment and Threshold Evaluation (GATE) methodology, illustrated in Figure \ref{fig:Methdology}. This methodology presents a systematic approach to optimizing QML models by using the GSI to identify and remove redundant gates in quantum circuits. This methodology follows a series of steps to create efficient and effective quantum models suited for practical applications. The models have been developed and tested in simulation environments, enabling a detailed evaluation of their performance and efficiency before deployment on real quantum hardware. %& PACO: QUITO ESTA FRASE DE MOMENTO PARA EVITAR QUE NOS O PIDAN: One of the main reasons for not executing the models on actual quantum computers is the limited accessibility to such hardware and the high costs associated with their usage.

Each stage of the methodology plays a critical role in transforming the raw data into an optimized quantum model. Next, we proceed to explain each step in detail:

\begin{figure}[!ht]
\centering
\includegraphics[width=\textwidth]{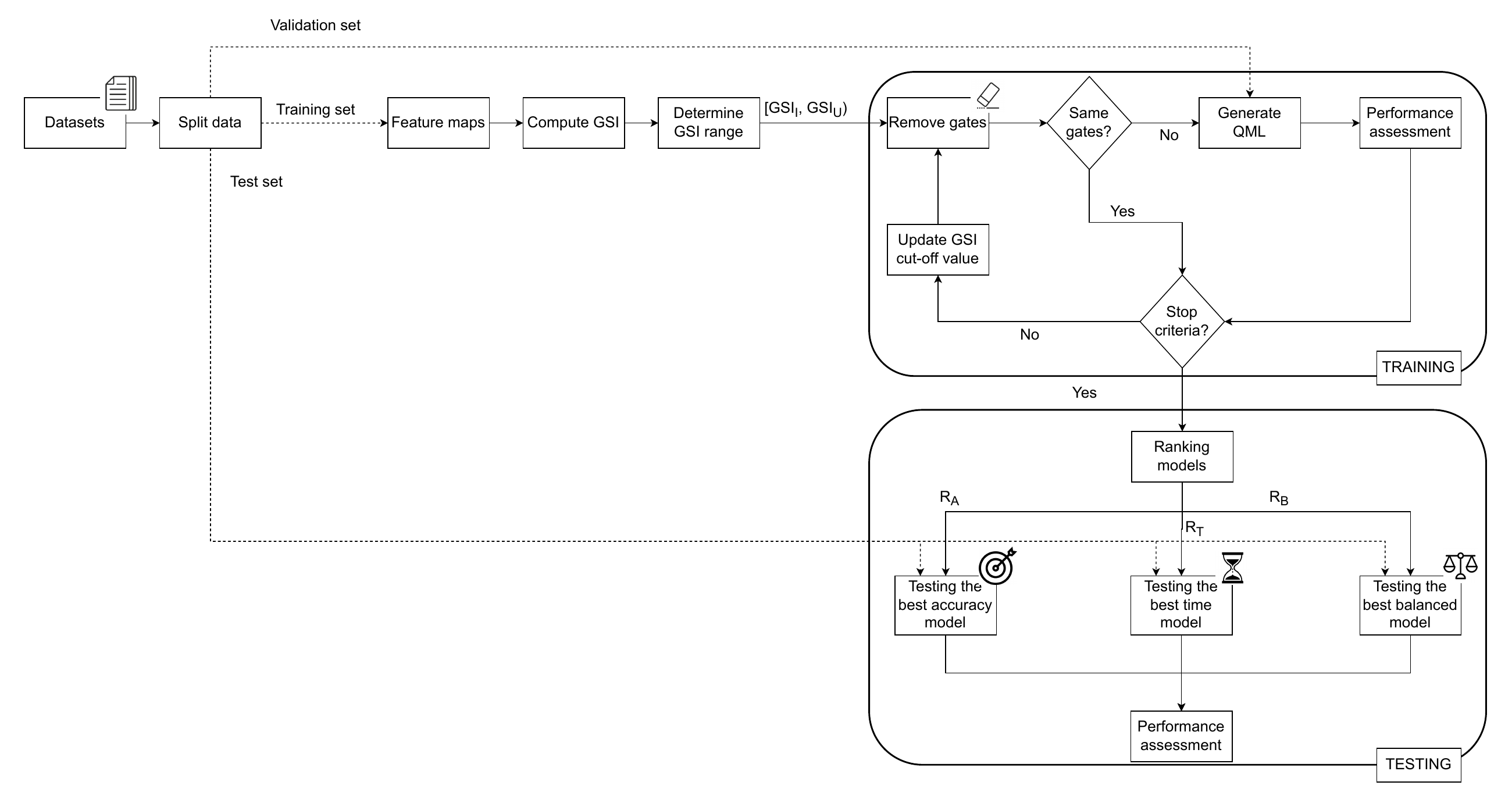}
\caption{Flowchart of the GATE methodology based on the GSI.}
\label{fig:Methdology}
\end{figure}

\subsubsection{Data preparation and splitting}
The process begins with the selection and preparation of datasets that serve as the basis for training, validation, and testing the quantum model. The dataset is divided into three parts: 60\% for training, 20\% for validation, and 20\% for testing. The training set is used to train the quantum model and develop feature mappings, which are transformations that encode classical data into quantum states. %By mapping data into a higher-dimensional quantum feature space, the model can uncover patterns or relationships that may be challenging to discern in the original data space, enhancing the model's ability to differentiate between classes.

Once the feature map is developed using the training set, the validation set is used to evaluate its effectiveness, ensuring that the model captures the essential features of the dataset without overfitting. Finally, the test set is used to provide an unbiased assessment of the model's generalization capability. %This systematic approach ensures the quantum model is both efficient and robust, capable of leveraging quantum resources to address complex real-world tasks.

\subsubsection{Feature mapping}
Once data have been split, the next step is to encode classical data into quantum states using feature-mapping techniques. In this methodology, the feature map is used for angle encoding, in which classical features are mapped to the rotation angles of quantum gates in a quantum circuit \cite{RATH24}. This type of encoding efficiently uses qubits and leverages quantum phenomena, such as superposition, to represent multiple data points simultaneously. This process preserves the structure and relationships in the classical data while enabling it to interact with quantum operations. The feature map also generates the quantum kernel, which is used to compute the kernel matrix and is a critical component in constructing the QML model. This foundational circuit will be later optimized in the following steps.

\subsubsection{Computing the GSI for each gate}
At this stage, the GSI is calculated for each gate in the quantum circuit, providing a numerical measure of its relevance based on the key metrics discussed in Section \ref{sec:GSI}. This assessment determines the contribution of each gate to the overall performance of the circuit.

\subsubsection{Determining the GSI iterating range}
Once the GSI values are computed for each gate, the next step is establishing a GSI range for iteration. This range is critical because it determines which gates will be retained and which will be considered for removal in each iteration.

Hence, the GSI range is obtained from a valid interval between the minimum and maximum GSI calculated values in the quantum circuit, where the upper bound, $GSI_u$, corresponds to the gate with the highest GSI value in the circuit and the lower bound, $GSI_l$, to the gate with the lowest one. During the training process, we iterate over the GSI interval to determine which gates remain and which are removed from the circuit at each iteration, setting the initial cut-off value to $GSI_l$ and the final cut-off value to $GSI_u$. Note that, in practice, the cut-off value is never set to $GSI_u$, as this could result in the removal of all gates. Therefore, the GSI iterating range is set to $[GSI_l, GSI_u)$, but ensuring that all qubits are active with at least one connected gate.

\subsubsection{Generation of the QML model}

The feature map plays a critical role in this process by encoding classical data in quantum states, serving as the foundation for building the QML model. At each iteration, a new QML is created by retaining the gates with GSI above the cut-off value.

This newly generated model represents a new version of the initial circuit, designed to reduce computational complexity while preserving accuracy and robustness.

\subsubsection{Performance assessment on the validation set}
The quantum model is then evaluated on the validation set. This step evaluates the model using metrics for accuracy and time. The validation process ensures that the model maintains adequate performance despite structural simplification. 

\subsubsection{Training stop criteria}
At each iteration, a new feature map is generated based on the gates that remain in the circuit.

The process has two stopping criteria. The first occurs when the circuit is evaluated after removing the gates whose GSI cutoff value is specified by $GSI_u$.

The second stopping criterion arises when, beyond certain GSI cut-off values, some qubits are left with no assigned gates. When this happens, the process of eliminating gates and generating new models is halted to ensure that the circuit remains functional and retains its computational capabilities.

At each iteration, the GSI cutoff value is updated by increasing it by an arbitrary step size, which depends on the specific problem and may vary with the associated GSI range. The process continues until one of the stopping criteria is met.

\subsubsection{Ranking of optimized models}
Once the iterative process has completed, the optimized circuit from each iteration is evaluated for accuracy and execution time. However, the best model has to be selected.

Thus, the models are ranked based on three key performance criteria: best accuracy model, best time model, and best-balanced model. The best accuracy model prioritizes the highest achievable accuracy, making it suitable for tasks where precision is critical. The best time model focuses on minimizing execution time, which is critical for applications that require rapid processing. However, when selecting the model with the shortest execution time, an additional condition is imposed: the chosen model's accuracy must be at least 15\% of the baseline model's accuracy. That is, an accuracy constraint is imposed, ensuring that circuits with optimal execution time still achieve high accuracy. Finally, the best-balanced model represents a trade-off between accuracy and time, offering a well-rounded option for scenarios where both factors are important. 
    
The method for computing these metrics is provided in Eqs. (\ref{eq:A}), (\ref{eq:T}) and (\ref{eq:B}): 

\begin{equation}\label{eq:A}
    A = \frac{TP + TN}{TP+TN+FP+FN}
\end{equation}
    
where $A$ identifies the accuracy achieved by the QML model with the evaluated quantum circuit, TP the true positives, TN the true negatives, FP the false positives, and FN the false negatives. 

\begin{equation}\label{eq:T}
    T = \text{execution time of QML model}
\end{equation}

where T identifies the execution time of the QML model for the evaluated quantum circuit.
    
\begin{equation}\label{eq:B}
    B = (A_n - A_b) + \frac{T_b - T_n}{T_b}
\end{equation}
    
where B identifies the balanced metric, $A_b$ and $T_b$ denote the accuracy and time of the baseline model (with all quantum gates included), and $A_n$ and $T_n$ represent the accuracy and execution time of a new model generated (by removing gates based on the GSI). Note that this metric may vary depending on the problem type (in this case, classification) and the specific metric used for optimization. On the one hand, T is an unbounded variable and, for this reason, the term evaluating the execution time must be normalized so that it falls in the range $[0,1]$. On the other hand, $A \in [0,1]$ and no normalization step must be carried out so that both terms equally contribute to B. 
    
Once these performance metrics are evaluated iteratively, the optimized circuits are ranked. In particular, three rankings are generated for each metric: $R_A$ for accuracy, $R_T$ for execution time, and $R_B$ for the balanced model. These rankings provide flexibility, allowing practitioners to choose a model based on specific performance needs.

\subsubsection{Final performance assessment}

The top-ranked models undergo a final performance assessment with the test set to validate their suitability for real-world applications. In particular, the PegasosQSVM performance with the optimized circuit is evaluated using, once again, metrics related to accuracy, execution time, or a balance between the two. This evaluation confirms that each model meets the required standards for accuracy and runtime.

\section{Results}
\label{sec:results}

The results of applying GATE to nine datasets are discussed in this section. Thus, the experimental setup is described in Section \ref{sec:experimental}. Section \ref{sec:data} describes each of the nine datasets used for training and testing. Finally, Section \ref{subsec:discussion} reports and discusses the results obtained by GATE when applied to such datasets.

\subsection{Experimental setup}\label{sec:experimental}

Two types of simulators were employed: a Free Noise Simulator (FNS), which utilizes Qiskit's default statevector simulator without any noise, and a Noise Simulator (NS), which incorporates a noise model derived from the real IBM backend (ibm\_brisbane) to replicate realistic quantum noise conditions.

The development environment used for this project was PyCharm (version 2023.2.5), which provided a robust and user-friendly platform for code development and debugging. The programming language used was Python, version 3.11.10, due to its compatibility with the quantum computing libraries employed. The simulators were run on a computer with the following specifications: an Apple M1 Pro processor, 16 GB of RAM, 1 TB of SSD storage, and Sequoia 15.0 as the operating system.

The framework used to construct the feature maps and calculate the GSI metric was Qiskit, version 1.3.4, developed by IBM. For model creation, training, and testing, the Qiskit Machine Learning module (version 0.8.2), maintained by the Qiskit community, was used. Additionally, the noise simulations were executed using Qiskit Aer, version 0.17.0, ensuring accurate modeling of quantum noise effects.

The feature map configuration used in the experiment was defined with specific parameters. The number of repetitions of the feature map circuit, denoted as Reps, was set to 1. Additionally, the entanglement configuration was defined as linear, establishing a linear connectivity pattern for entangling the qubits.

The PegasosQSVM implementation was configured with specific parameters. The parameter "C" was set to 5000, controlling the balance between minimizing the training error and maintaining model complexity. Additionally, the parameter "num\_steps" was set to 500, specifying the number of steps for the Pegasos algorithm during optimization.

The QNN implementation was configured with specific hyperparameters. The variational ansatz was defined with reps = $1$. Training was carried out using the COBYLA optimizer, with maxiter set to $100$ and tol set to $10^{-4}$. In addition, the trainable parameters were initialized randomly from a uniform distribution in the interval $[0, 2\pi]$.

The quantum hardware experiments were carried out on the IBM ibm\_strasbourg processor. This device belongs to the Eagle r3 family and comprises 127 superconducting qubits arranged in a hexagonal configuration, providing efficient connectivity between qubits. According to the closest available calibration, the average relaxation and coherence times were $T_1 = 264.52\,\mu s$ and $T_2 = 123.8\,\mu s$, respectively. The median readout error was $2.60\times 10^{-2}$, while the median error rates for the native single-qubit ($sx$) and two-qubit ($ecr$) gates were $2.441\times 10^{-4}$ and $7.51\times 10^{-3}$, respectively. The device operates with the native gate set {ecr, id, rz, sx, x}, and its performance, measured in circuit layer operations per second (CLOPS), was approximately $250K$.

\subsubsection{Execution environments}\label{sec:foundationsScalability}

This section presents the execution environments considered in this research for computing the quantities required by GSI. Since these environments rely on different state representations and execution strategies, they exhibit different computational costs and scalability behavior. In particular, we consider three classical simulation-based approaches, namely DM, MPS, and TN, together with a real device based approach (RD). Their empirical scalability is analyzed in Section~\ref{sec:scalabilityResults}.

\paragraph{Density Matrix method}

For an $n$-qubit pure state $|\psi\rangle$, the corresponding density matrix is given by Eq. (\ref{eq:densityMatrixEq}).

\begin{equation}\label{eq:densityMatrixEq}
    \rho = |\psi\rangle\langle\psi|.
\end{equation}

This is a $2^n \times 2^n$ matrix. To obtain the reduced density matrix of a subsystem $A$ (e.g., a single qubit or a subset of qubits), one performs a partial trace over the complementary subsystem $B$, as shown in Eq. (\ref{eq:traceEq}).

\begin{equation}\label{eq:traceEq}
    \rho_A = \mathrm{Tr}_B(\rho).
\end{equation}

The entanglement contribution can then be quantified through the von Neumann entropy of the reduced state, which was defined in Eq. (\ref{eq:entaglement}).

The entanglement contribution can then be quantified through the von Neumann entropy of the reduced state, as defined in Eq.~(\ref{eq:entaglement}), which satisfies $0 \le S(\rho_A) \le \log_2(d_A)$, where $d_A$ is the Hilbert-space dimension of subsystem $A$ (e.g., $d_A=2$ for a single qubit). In dense density-matrix simulation, storing $\rho$ already requires memory that grows as $O(4^n)$, and operations needed for entropy evaluation (e.g., diagonalization of $\rho_A$) scale accordingly, which makes this approach impractical for large $n$.

\paragraph{Matrix Product State method}

To mitigate the exponential overhead of dense simulation, we also consider tensor-network techniques, specifically the MPS representation. In the MPS formalism, an $n$-qubit state can be expressed as a product of local tensors, as shown in Eq. (\ref{eq:tensorsEq}).

\begin{equation}\label{eq:tensorsEq}
|\psi\rangle = \sum_{i_1, i_2, \dots, i_n} A^{[1]}_{i_1}\, A^{[2]}_{i_2}\, \cdots\, A^{[n]}_{i_n}\, |i_1 i_2 \cdots i_n\rangle.
\end{equation}

Each tensor $A^{[k]}_{i_k}$ includes bond indices that capture correlations between adjacent qubits. The maximum bond dimension, denoted by $\chi$, controls how much entanglement the representation can faithfully capture. When entanglement remains moderate, a relatively small $\chi$ suffices, and the cost of updating and contracting the network scales polynomially in $n$ and $\chi$ (commonly reported as $O(n\chi^3)$ up to method-dependent constants), representing a major improvement over the dense approach.

In practice, reduced density matrices required by the GSI components can be obtained by contracting the MPS and tracing out the qubits that are not of interest. For example, if the entanglement proxy is computed from a single qubit, the reduced density matrix is only $2\times 2$, and its eigenvalues (and entropy) can be computed efficiently. The MPS method may introduce an approximation when truncation is enabled (through a truncation threshold and/or a maximum bond dimension) to keep the simulation tractable. Increasing $\chi$ or tightening the truncation threshold improves accuracy but increases runtime and memory usage. In the worst case, if the circuit generates near-maximal entanglement, the required bond dimension may grow rapidly, and the complexity can approach exponential behavior.

\paragraph{Tensor Network method}

TN simulation represents the circuit as a network of small tensors (one per gate, initial state, and measurement) connected according to the circuit wiring, and calculates the necessary quantities by contracting this network. Unlike dense simulation, TN methods do not explicitly construct the full $2^n$-dimensional state vector or the $2^n \times 2^n$ density matrix. Instead, the computational cost is dominated by the contraction order (also called the contraction path). Conceptually, a TN contraction can be written as a sequence of index summations, as shown in Eq. (\ref{eq:tn_contraction}).

\begin{equation}\label{eq:tn_contraction}
\mathcal{A} \;=\; \sum_{\{i\}} \prod_{k} T^{(k)}_{\{i\}_k},
\end{equation}

where each $T^{(k)}$ is a local tensor (associated with gates or state preparation) and the sum runs over all internal indices $\{i\}$ that connect the tensors. In general, the time and memory required to contract a tensor network scale exponentially in the tree width of the network (or, equivalently, the size of the largest intermediate tensor created along the chosen contraction path). This dependence is often summarized as depicted in Eq. (\ref{eq:tn_treewidth_scaling})

\begin{equation}\label{eq:tn_treewidth_scaling}
\text{cost} \;=\; O\!\left(\mathrm{poly}(N)\, 2^{w}\right),
\end{equation}

where $N$ is the number of tensors (typically proportional to circuit size) and $w$ denotes the effective tree width induced by the contraction ordering. For structured circuits with sparse connectivity (e.g., entanglement patterns close to 1D), the effective tree width can remain small and contractions can be performed efficiently. However, for highly interleaved circuits with dense connectivity, the tree width often grows rapidly, resulting in large intermediate tensors and a sharp increase in runtime and memory, which can lead to out-of-memory failures.

In practice, TN simulators rely on heuristic route search and may also apply techniques such as slicing (index splitting to reduce memory peak) to trade memory for additional computation. For our GSI workflow, TN methods are attractive because they can provide reduced-density matrices (or local marginals) without explicit storage of the full state. However, their performance is sensitive to the circuit structure and the overhead introduced by repeated extraction of intermediate reduced states throughout the circuit evolution.

\paragraph{Real hardware method}

In the actual hardware, GSI components are obtained exclusively from measurement statistics, without the need to access the complete quantum state. This makes the hardware formulation the most scalable of all the approaches considered, since its cost scales primarily with the number of circuits and shots executed, rather than with the dimension of the Hilbert space. Although this formulation is statistical, since each metric is estimated from counts of finite shot measurements, it is still consistent with the definitions of the simulator/emulator, since in the noise-free limit and with a sufficient number of shots, the superposition-based estimator for $\widehat{F}$ reproduces the corresponding amount of state superposition, and the single-qubit entropy reconstructed from Pauli expectations coincides with the reduced state entropy used as a proxy for entanglement. 

Consequently, the hardware procedure produces the same general trends and gate classification conclusions as simulation/emulation methods in comparable environments, while also being the most relevant approach from a practical standpoint, as it is the only one that operates directly on a physical device and captures the effects that ultimately matter in real executions (noise, decoherence, and readout errors). The entire formulation was presented in Section \ref{sec:GSIRH}.

\subsubsection{Datasets description}\label{sec:data}
This section describes the datasets used to test the effectiveness of the proposed methodology. Table \ref{tab:datasets} presents the reference, name, number of instances, number of features, and number of classes for each dataset.

The BreastW, Corral, Glass2, Monk, Flare, Vote, and Saheart datasets are part of the Penn Machine Learning Benchmarks (PMLB) suite \cite{Olson2017PMLB}, which provides a standardized collection of datasets for evaluating and comparing machine learning algorithms. These datasets encompass a range of binary classification tasks, from medical diagnosis to material classification, with varying levels of complexity.

The BreastW dataset, also known as Breast Cancer Wisconsin, is used to classify tumors as benign or malignant based on 10 attributes. Corral, Glass2, and Monk are smaller datasets designed for binary classification, each with different levels of feature interaction. The Flare dataset is used to predict solar flare occurrences using 11 attributes, whereas the Vote dataset classifies voting patterns of U.S. Congress members using 17 features. Finally, the Saheart dataset contains clinical data to predict the likelihood of heart disease.

In addition, other datasets have been included to test the methodology in different domains. The Heart dataset \cite{fedesoriano2021heart}, derived from patient records, aims to predict the presence of heart disease based on 12 clinical attributes. The Fitness dataset \cite{dosad2024fitness} analyzes membership retention in a fitness club, using 7 features to determine whether a customer is likely to remain a member. 

Notably, all datasets are binary classification tasks, as the models in the Qiskit library, which will be used in this study, are specifically designed for binary classification.

\begin{table}[htb]
\centering
\caption{Description of the datasets.}
\label{tab:datasets}
\setlength{\tabcolsep}{6pt}
\renewcommand{\arraystretch}{1.2}
\begin{tabular}{|c|c|c|c|c|}
\hline
\textbf{Ref.} & 
\textbf{Name} & 
\textbf{Instances} & 
\textbf{Features} & 
\textbf{Classes} \\
\hline
\cite{Olson2017PMLB}         & BreastW     & 699     & 10 & 2 \\
\cite{Olson2017PMLB}         & Corral      & 160     & 7  & 2 \\
\cite{dosad2024fitness}      & Fitness     & 1500    & 7  & 2 \\
\cite{Olson2017PMLB}         & Glass2      & 163     & 10 & 2 \\
\cite{fedesoriano2021heart}  & Heart       & 918     & 12 & 2 \\
\cite{Olson2017PMLB}         & Monk        & 556     & 7  & 2 \\
\cite{Olson2017PMLB}         & Flare       & 1066    & 11 & 2 \\
\cite{Olson2017PMLB}         & Vote        & 435     & 17 & 2 \\
\cite{Olson2017PMLB}         & Saheart     & 462     & 10 & 2 \\
\hline
\end{tabular}
\end{table}

\subsection{Discussion} \label{subsec:discussion} 

This section presents the results obtained after applying GATE. Once the feature maps have been created, the GSI is calculated for each gate. Table \ref{tab:timeExecution} shows the execution time required to calculate the GSI across all datasets and methods used for this calculation. 

A general trend observed in the updated results is that the GSI computation time depends more on the execution backend than on the specific dataset. Specifically, the MPS backend exhibits the most stable behavior, with a nearly constant execution time of 0.01 seconds in all cases. The TN backend is relatively consistent, with response times primarily ranging from 1.7 to 2.7 seconds, whereas Vote stands out as the main exception at 4.15 seconds. In contrast, RD consistently requires longer times, generally between 6 and 8.5 seconds, and again, Vote reaches the highest value at 10.00 seconds. The DM backend shows the greatest variability, ranging from very short times on several datasets to significantly longer times in cases such as Heart and Saheart.

These results indicate that the choice of backend is the primary factor determining the computational cost of the GSI. MPS offers the best scalability, whereas TN provides a balanced intermediate solution with low, stable execution times. RD, although slower, remains within practical limits, which supports the method’s applicability beyond simulation. It should be noted that these times also depend on the circuit construction, specifically the number of repetitions and the design of the interleaving, since more complex interleaving schemes reduce scalability. All of this is better illustrated in the stability analysis presented in Section \ref{sec:resultsScalability}.

Finally, the fact that only a few datasets, particularly “Vote”, behave as outliers suggests that the specific complexity of each dataset has some influence, but much less than the computational configuration itself.

\begin{table}[htb]
\centering
\caption{GSI computation time for each dataset.}
\label{tab:timeExecution}
\setlength{\tabcolsep}{12pt}
\renewcommand{\arraystretch}{1.2}
\begin{tabular}{|l|c|c|c|c|}
\hline
\textbf{Dataset} & \textbf{DM time (s)} & \textbf{MPS time (s)} & \textbf{TN time (s)} & \textbf{RD time (s)} \\
\hline
BreastW   & 0.04 & 0.01 & 2.53 & 6.14 \\
Corral    & 0.01 & 0.01 & 1.81 & 6.35 \\
Fitness   & 0.01 & 0.01 & 1.78 & 6.35 \\
Glass2    & 0.10 & 0.01 & 2.42 & 8.45 \\
Heart     & 2.34 & 0.01 & 2.65 & 6.50 \\
Monk      & 0.01 & 0.01 & 1.72 & 6.35 \\
Flare     & 0.12 & 0.01 & 2.44 & 6.03 \\
Vote      & - & 0.01 & 4.15 & 10.00 \\
Saheart   & 0.53 & 0.01 & 2.38 & 6.14 \\
\hline
\end{tabular}
\end{table}

The results obtained from the training process are presented below. Each of these tables contains six columns. The first column indicates the dataset, and the second shows the GSI index value used for gate selection and model generation. The third column shows the number of gates. The next three columns present the accuracy, runtime, and ranking of the best model for accuracy (A), time (T), and balance (B). In this case, there are two tables for the models trained on the ideal simulation: one containing the first 5 datasets and another containing the next 4. This was done to improve data visualization. The same will apply to models trained in an environment with emulated noise. Finally, a table will be presented for the models trained in a real-world environment; as shown, there are only 3. This is due to practical limitations associated with the use of real quantum hardware, particularly in terms of resource availability, runtime, and computational cost, which necessitate restricting the number of experiments conducted in this scenario.

\subsubsection{Discussion of PegasusQSVM results}

The results obtained for PegasusQSVM in the simulator (Table \ref{tab:training_fns_1_dedup_QSVM} and Table \ref{tab:training_fns_2_dedup_QSVM}) show a clear general trend, as the GSI value increases, the number of gates decreases and, in most cases, execution time is also reduced. At the same time, classification performance is maintained or even improved. This behavior suggests that the original models are not always the most efficient configurations, as moderate simplification of the circuit typically yields better overall results. Across several datasets, the configurations ranked highest by $R_{ATB}$ are not found at the highest GSI values, but rather at intermediate points, indicating that an appropriate balance between circuit simplification and predictive power is required.

A second notable trend is that the effect of GSI is not uniform across all datasets. In many cases, such as BreastW, Heart, Vote, Monk, and Corral, the simplified models outperform the baseline model not only in runtime but also in accuracy, demonstrating that reducing circuit complexity can improve the model’s expressive efficiency. However, the results also show that excessive simplification can be detrimental. This is particularly evident in datasets where higher GSI values result in a clear drop in accuracy, even as runtime continues to decrease. Therefore, reducing the number of gates cannot be interpreted as a purely monotonic improvement, but rather as a trade-off in which the smaller circuit is not necessarily the most suitable.

\begin{table}[!htb]
\caption{Training results (PegasusQSVM) for the first five datasets in Simulator.}
\label{tab:training_fns_1_dedup_QSVM}
\setlength{\tabcolsep}{4pt}
\renewcommand{\arraystretch}{1.2}
\centering
\begin{tabular}{|l|cc|ccc|}
\hline
\textbf{Dataset} & \textbf{GSI} & \textbf{\#Gates} & \textbf{Acc.} & \textbf{Time} & \boldmath{$R_{ATB}$} \\
\hline
\multirow[c]{4}{*}{BreastW} & 0.518 & 42 & 0.792 & 187 & -- \\
                            & 0.538 & 33 & 0.785 & 144 & 2-2-3 \\
                            & 0.558 & 29 & 0.892 & 116 & \textbf{1-1-1} \\
                            & 0.578 & 28 & 0.628 & 103 & 3-3-2 \\
\hline
\multirow[c]{8}{*}{Heart}   & 0.533 & 52 & 0.717 & 315 & -- \\
                            & 0.573 & 48 & 0.771 & 325 & 2-4-6 \\
                            & 0.593 & 47 & 0.760 & 285 & 3-3-4 \\
                            & 0.613 & 45 & 0.733 & 258 & 4-2-2 \\
                            & 0.633 & 43 & 0.788 & 237 & \textbf{1-1-1} \\
                            & 0.653 & 34 & 0.576 & 217 & 5-5-3 \\
                            & 0.673 & 31 & 0.576 & 321 & 6-6-7 \\
                            & 0.693 & 24 & 0.576 & 253 & 7-7-5 \\
\hline
\multirow[c]{4}{*}{Flare}   & 0.520 & 47 & 0.591 & 213 & -- \\
                            & 0.545 & 40 & 0.845 & 172 & \textbf{1}-3-2 \\
                            & 0.570 & 35 & 0.830 & 158 & 2-2-\textbf{1} \\
                            & 0.595 & 32 & 0.737 & 150 & 3-\textbf{1}-3 \\
\hline
\multirow[c]{3}{*}{Vote}    & 0.625 & 77 & 0.747 & 485 & -- \\
                            & 0.700 & 75 & 0.793 & 379 & 2-2-2 \\
                            & 0.725 & 69 & 0.931 & 353 & \textbf{1-1-1} \\
\hline
\multirow[c]{5}{*}{Glass2}  & 0.500 & 42 & 0.727 & 106 & -- \\
                            & 0.520 & 37 & 0.696 & 95  & 4-3-4 \\
                            & 0.540 & 35 & 0.727 & 95  & 3-4-2 \\
                            & 0.560 & 32 & 0.787 & 88  & \textbf{1}-2-3 \\
                            & 0.600 & 30 & 0.787 & 86  & 2-\textbf{1-1} \\
\hline
\end{tabular}
\end{table}

\begin{table}[!htb]
\caption{Training results (PegasusQSVM) for the last four datasets in the simulator.}
\label{tab:training_fns_2_dedup_QSVM}
\setlength{\tabcolsep}{4pt}
\renewcommand{\arraystretch}{1.2}
\centering
\begin{tabular}{|l|cc|ccc|}
\hline
\textbf{Dataset} & \textbf{GSI} & \textbf{\#Gates} & \textbf{Acc.} & \textbf{Time} & \boldmath{$R_{ATB}$} \\
\hline
\multirow[c]{5}{*}{Fitness}  & 0.520 & 27 & 0.336 & 159  & -- \\
                             & 0.540 & 26 & 0.670 & 153  & \textbf{1}-4-4 \\
                             & 0.560 & 21 & 0.653 & 148  & 2-3-3 \\
                             & 0.580 & 18 & 0.643 & 120  & 3-2-2 \\
                             & 0.620 & 16 & 0.536 & 114  & 4-\textbf{1}-\textbf{1} \\
\hline
\multirow[c]{3}{*}{Monk}     & 0.543 & 26 & 0.675 & 75   & -- \\
                             & 0.563 & 23 & 0.639 & 66   & 2-2-2 \\
                             & 0.603 & 18 & 0.783 & 51   & \textbf{1-1-1} \\
\hline
\multirow[c]{5}{*}{Saheart}  & 0.531 & 42 & 0.645 & 561  & -- \\
                             & 0.556 & 33 & 0.536 & 416  & 3-4-4 \\
                             & 0.581 & 29 & 0.462 & 302  & 4-3-3 \\
                             & 0.606 & 27 & 0.645 & 248  & \textbf{1}-\textbf{1}-\textbf{1} \\
                             & 0.631 & 22 & 0.623 & 262  & 2-2-2 \\
\hline
\multirow[c]{5}{*}{Corral}   & 0.566 & 26 & 0.656 & 21   & -- \\
                             & 0.605 & 24 & 0.968 & 18   & \textbf{1}-\textbf{1}-\textbf{1} \\
                             & 0.645 & 20 & 0.812 & 23   & 2-3-3 \\
                             & 0.665 & 19 & 0.812 & 22   & 3-2-2 \\
                             & 0.685 & 14 & 0.531 & 33   & 4-4-4 \\
\hline
\end{tabular}
\end{table}

In the emulator environment (Table \ref{tab:training_ns_1_dedup_QSVM} and Table \ref{tab:training_ns_2_dedup_QSVM}), the results show that the presence of noise makes the response to GSI adjustments more irregular than in the ideal simulator. Although reductions in the number of gates are still observed as GSI increases, the relationship between circuit simplification and predictive performance becomes less uniform and more dependent on the dataset. In this scenario, improvements are still observed in many cases, but they are less uniform, suggesting that noise introduces an additional source of sensitivity in the model selection process. As a result, the best configuration is more clearly associated with a balanced operating point than with a simple trend toward smaller circuits.

A distinctive feature of the emulator results is that runtime becomes much more significant in the models' overall behavior. The computational cost is substantially higher than in the ideal scenario, especially on datasets such as Fitness and Vote, making the efficiency dimension more relevant when comparing candidate configurations. Under these conditions, the ranking provided by $R_{ATB}$ is particularly informative, as it highlights solutions that remain competitive in terms of accuracy while avoiding unnecessarily costly circuits. This is especially important because the configuration with the highest accuracy is not always the most attractive when execution cost under noise is considered.

Another important observation is that the emulator results reveal a more pronounced separation between stable and unstable datasets. Some problems continue to benefit significantly from intermediate GSI values, whereas others exhibit more pronounced performance declines when circuit complexity is reduced beyond a certain point. This indicates that, in noisy environments, the method not only adapts the circuit’s structural size but also highlights each dataset’s sensitivity to reduced complexity under non-ideal conditions. Therefore, the emulator configuration provides a more demanding validation scenario, in which the utility of GSI lies not only in producing smaller models but also in identifying configurations that remain reliable when noise is introduced into the training process.

\begin{table}[!htb]
\caption{Training results (PegasusQSVM) for the first five datasets in the emulator.}
\label{tab:training_ns_1_dedup_QSVM}
\setlength{\tabcolsep}{4pt}
\renewcommand{\arraystretch}{1.2}
\centering
\begin{tabular}{|l|cc|ccc|}
\hline
\textbf{Dataset} & \textbf{GSI} & \textbf{\#Gates} & \textbf{Acc.} & \textbf{Time} & \boldmath{$R_{ATB}$} \\
\hline
\multirow[c]{4}{*}{BreastW} & 0.518 & 42 & 0.828 & 1990 & -- \\
                            & 0.538 & 33 & 0.728 & 1651 & 2-\textbf{1}-2 \\
                            & 0.558 & 29 & 0.878 & 1796 & \textbf{1}-2-\textbf{1} \\
                            & 0.578 & 28 & 0.628 & 1720 & 3-3-3 \\
\hline
\multirow[c]{8}{*}{Heart}   & 0.533 & 52 & 0.728 & 4236 & -- \\
                            & 0.553 & 48 & 0.663 & 4131 & 4-3-6 \\
                            & 0.593 & 47 & 0.755 & 4040 & 2-2-2 \\
                            & 0.613 & 45 & 0.739 & 4258 & 3-4-5 \\
                            & 0.633 & 43 & 0.798 & 3920 & \textbf{1}-\textbf{1}-\textbf{1} \\
                            & 0.653 & 34 & 0.576 & 3497 & 5-5-3 \\
                            & 0.673 & 31 & 0.467 & 3575 & 7-7-7 \\
                            & 0.693 & 24 & 0.576 & 3534 & 6-6-4 \\
\hline
\multirow[c]{4}{*}{Flare}   & 0.520 & 47 & 0.605 & 2244 & -- \\
                            & 0.545 & 40 & 0.840 & 2100 & \textbf{1}-3-3 \\
                            & 0.570 & 35 & 0.835 & 1786 & 2-\textbf{1}-\textbf{1} \\
                            & 0.595 & 32 & 0.746 & 1837 & 3-2-2 \\
\hline
\multirow[c]{3}{*}{Vote}    & 0.625 & 77 & 0.908 & 10801 & -- \\
                            & 0.700 & 75 & 0.942 & 9156  & \textbf{1}-\textbf{1}-\textbf{1} \\
                            & 0.725 & 69 & 0.917 & 13564 & 2-2-2 \\
\hline
\multirow[c]{5}{*}{Glass2}  & 0.500 & 42 & 0.696 & 1530 & -- \\
                            & 0.520 & 37 & 0.727 & 1426 & 3-4-4 \\
                            & 0.540 & 35 & 0.757 & 1413 & 2-3-2 \\
                            & 0.580 & 32 & 0.727 & 1394 & 4-2-3 \\
                            & 0.600 & 30 & 0.818 & 1364 & \textbf{1}-\textbf{1}-\textbf{1} \\
\hline
\end{tabular}
\end{table}

\begin{table}[!htb]
\caption{Training results (PegasusQSVM) for the last four datasets in the emulator.}
\label{tab:training_ns_2_dedup_QSVM}
\setlength{\tabcolsep}{4pt}
\renewcommand{\arraystretch}{1.2}
\centering
\begin{tabular}{|l|cc|ccc|}
\hline
\textbf{Dataset} & \textbf{GSI} & \textbf{\#Gates} & \textbf{Acc.} & \textbf{Time} & \boldmath{$R_{ATB}$} \\
\hline
\multirow[c]{5}{*}{Fitness}  & 0.520 & 27 & 0.316 & 11984  & -- \\
                             & 0.540 & 26 & 0.686 & 12536  & 2-4-4 \\
                             & 0.560 & 21 & 0.676 & 7704   & 3-3-3 \\
                             & 0.580 & 18 & 0.690 & 6365   & \textbf{1}-2-\textbf{1} \\
                             & 0.620 & 16 & 0.560 & 5191   & 4-\textbf{1}-2 \\
\hline
\multirow[c]{3}{*}{Monk}     & 0.543 & 26 & 0.693 & 4864   & -- \\
                             & 0.563 & 23 & 0.600 & 2965   & 2-2-2 \\
                             & 0.583 & 18 & 0.783 & 2715   & \textbf{1-1-1} \\
\hline
\multirow[c]{5}{*}{Saheart}  & 0.531 & 42 & 0.494 & 2138   & -- \\
                             & 0.556 & 33 & 0.612 & 2062   & 2-4-4 \\
                             & 0.581 & 29 & 0.645 & 1641   & \textbf{1}-2-2 \\
                             & 0.606 & 27 & 0.430 & 1685   & 4-3-3 \\
                             & 0.631 & 22 & 0.602 & 1441   & 3-\textbf{1}-\textbf{1} \\
\hline
\multirow[c]{5}{*}{Corral}   & 0.566 & 26 & 0.906 & 1439   & -- \\
                             & 0.585 & 24 & 1.000 & 1227   & \textbf{1}-\textbf{1}-\textbf{1} \\
                             & 0.645 & 20 & 0.875 & 1715   & 2-3-2 \\
                             & 0.665 & 19 & 0.812 & 1660   & 3-2-3 \\
                             & 0.685 & 14 & 0.531 & 1362   & 4-4-4 \\
\hline
\end{tabular}
\end{table}

The results obtained on real hardware (Table \ref{tab:training_rd_2_QSVM}) show that the GSI effect remains beneficial, although the behavior is less consistent than in simulator and emulator environments. In this scenario, reducing the number of gates is still generally associated with shorter execution times, but the improvements in accuracy are more selective. It should also be noted that, in this case, a noise mitigation technique based on dynamic decoupling was incorporated, as mentioned in Section \ref{sec1}. Even with this additional support, the results are still more affected by device-level instability than in the previous environments.

A notable trend is that the top-performing configurations are, once again, not found in the baseline configuration, but rather in intermediate or moderately simplified circuits. Across the three datasets evaluated, these configurations offer a better balance between predictive performance and execution cost than the original models. At the same time, the results show that fewer logic gates do not necessarily imply better performance, as some reduced circuits primarily improve efficiency without achieving maximum accuracy.

Although the analysis on actual hardware is limited to only three datasets, the observed behavior is consistent with previous findings. The baseline is not necessarily the most suitable option, and controlled reductions in circuit complexity can still obtain more competitive and balanced models under real-world execution conditions.

\begin{table}[!htb]
\caption{Training results (PegasusQSVM) for three datasets in real hardware.}
\label{tab:training_rd_2_QSVM}
\setlength{\tabcolsep}{4pt}
\renewcommand{\arraystretch}{1.2}
\centering
\begin{tabular}{|l|cc|ccc|}
\hline
\textbf{Dataset} & \textbf{GSI} & \textbf{\#Gates} & \textbf{Acc.} & \textbf{Time} & \boldmath{$R_{ATB}$} \\
\hline
\multirow[c]{4}{*}{Corral}   & 0.583 & 27 & 0.593 & 175 & -- \\
                             & 0.603 & 26 & 0.593 & 94 & 3-\textbf{1}-3 \\
                             & 0.623 & 25 & 0.750 & 103 & \textbf{1}-3-\textbf{1} \\
                             & 0.703 & 24 & 0.625 & 95 & 2-2-2 \\
\hline
\multirow[c]{9}{*}{glass2}   & 0.384 & 42 & 0.393 & 98  & -- \\
                             & 0.404 & 39 & 0.393 & 96  & 3-3-3 \\
                             & 0.424 & 38 & 0.393 & 92  & 4-2-2 \\
                             & 0.444 & 37 & 0.393 & 101 & 5-4-4 \\
                             & 0.464 & 36 & 0.393 & 150 & 6-8-8 \\
                             & 0.484 & 33 & 0.393 & 107 & 7-6-6 \\
                             & 0.504 & 31 & 0.454 & 135 & 2-7-7 \\
                             & 0.544 & 30 & 0.393 & 106 & 8-5-5 \\
                             & 0.564 & 29 & 0.545 & 88  & \textbf{1}-\textbf{1}-\textbf{1} \\
\hline
\multirow[c]{4}{*}{monk}     & 0.466 & 27 & 0.540 & 373 & -- \\
                             & 0.566 & 25 & 0.639 & 315 & \textbf{1}-3-3 \\
                             & 0.666 & 23 & 0.621 & 215 & 3-\textbf{1}-2 \\
                             & 0.686 & 22 & 0.630 & 217 & 2-2-\textbf{1} \\
\hline
\end{tabular}
\end{table}

\begin{figure}[!ht]
\centering
\includegraphics[width=\textwidth]{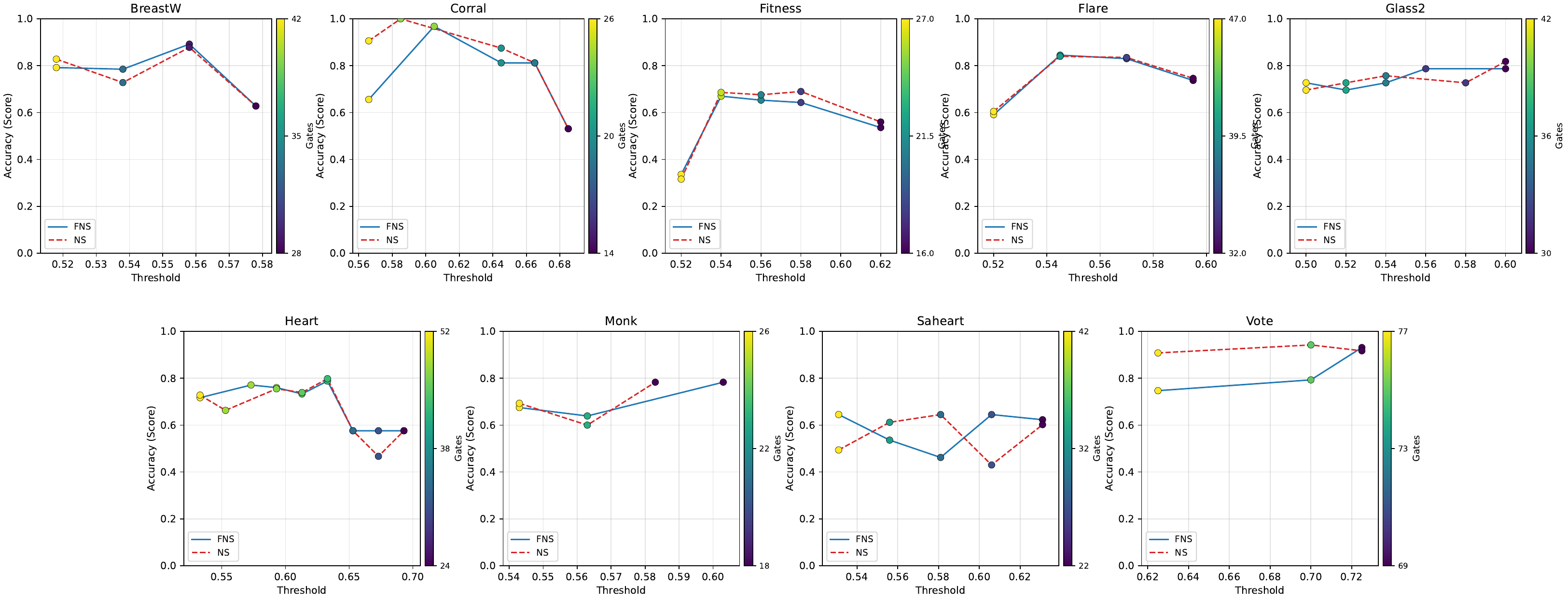}
\caption{Accuracy obtained for the different thresholds of GSI for all datasets with noise and without noise during the training phase (PegasusQSVM).}
\label{fig:ThresVSAcc_QSVM}
\end{figure}

The figure \ref{fig:ThresVSAcc_QSVM} complements the previous tables by showing more clearly how accuracy evolves as the threshold increases, both in the ideal simulator and in the noisy emulator. In this plot, the point farthest to the left corresponds to the reference model, from which subsequent configurations progressively modify the circuit by increasing the threshold value. In general, the trajectories show that the best results are not typically found at the lowest or highest threshold values, but rather at intermediate points, where a better balance is achieved between circuit reduction and predictive performance. This pattern is observed across several datasets, although the exact position of the most favorable threshold varies with the problem.

The comparison between FNS and NS also shows that noise does not completely alter the overall behavior of the models, as both curves generally retain a similar overall shape; however, it does make the evolution less stable and more irregular. In some datasets, the two fits remain close, whereas in others the noisy case exhibits more pronounced fluctuations or steeper drops at certain thresholds. Therefore, the figure reinforces the idea that threshold tuning should be understood as a tuning mechanism whose most useful values are typically found in an intermediate range, whereas its final effect depends on each dataset’s sensitivity to noise and the circuit's complexity.

\begin{figure}[!ht]
\centering
\includegraphics[width=\textwidth]{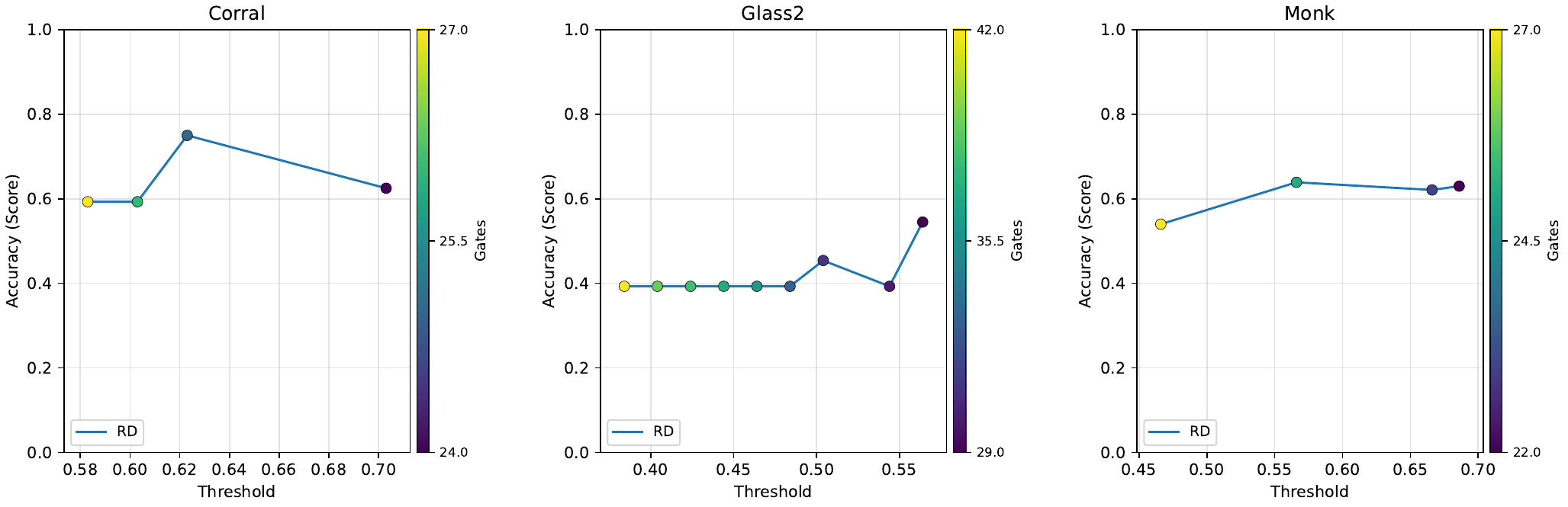}
\caption{Accuracy obtained for the different thresholds of GSI for three datasets in Real Device during the training phase (PegasusQSVM).}
\label{fig:ThresVSAccRD_QSVM}
\end{figure}

The real hardware figure \ref{fig:ThresVSAccRD_QSVM} provides a concise visual overview of how accuracy varies with the threshold in the RD environment. In general, the best results are not concentrated around the baseline model either, but rather at intermediate threshold values, demonstrating that moderate circuit simplification can continue to improve the balance between performance and efficiency under real-world execution conditions. At the same time, the trajectories are less smooth and more irregular than in simulation-based settings, consistent with the greater influence of hardware noise and execution variability. This effect is particularly evident in the differing responses observed in Corral, Glass2, and Monk, confirming that the most appropriate threshold still depends on the dataset, even when only a limited number of experiments can be conducted with real devices.

\begin{figure}[!ht]
\centering
\includegraphics[width=\textwidth]{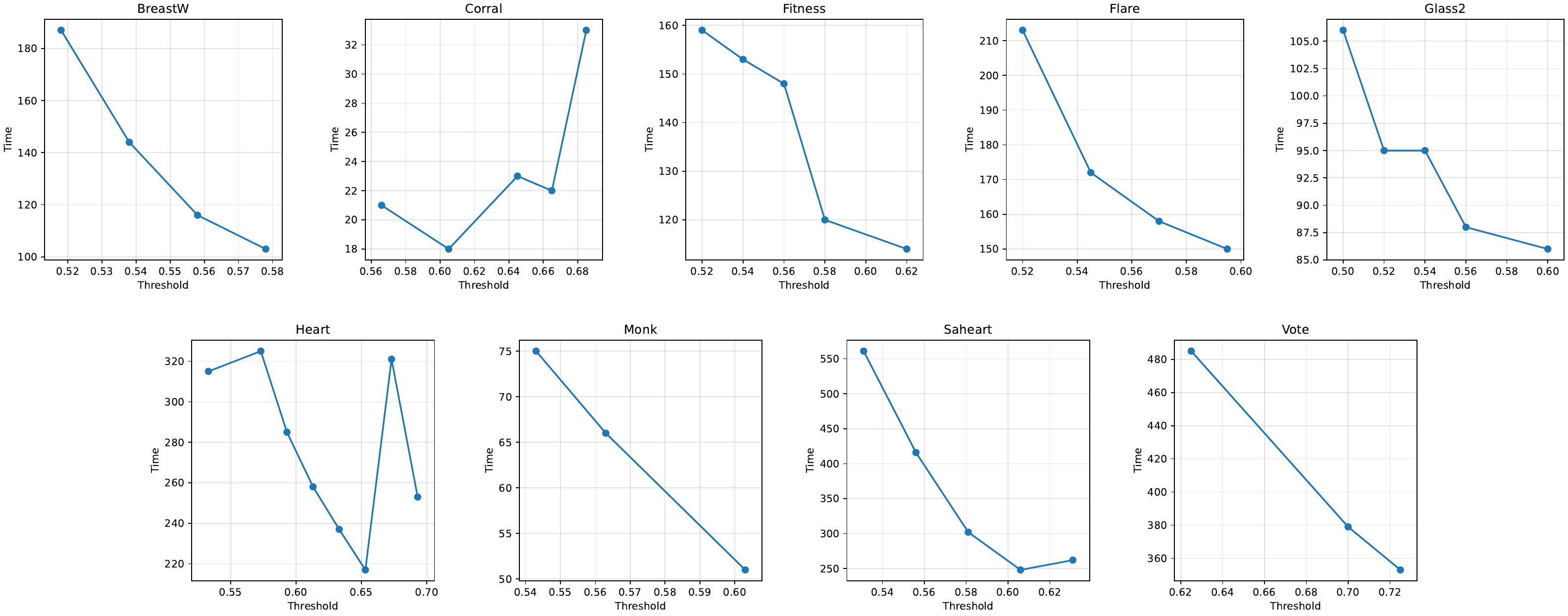}
\caption{Time obtained for the different thresholds of GSI for all datasets without noise during the training phase (PegasusQSVM).}
\label{fig:ThresVSTimeFNS_QSVM}
\end{figure}

\begin{figure}[!ht]
\centering
\includegraphics[width=\textwidth]{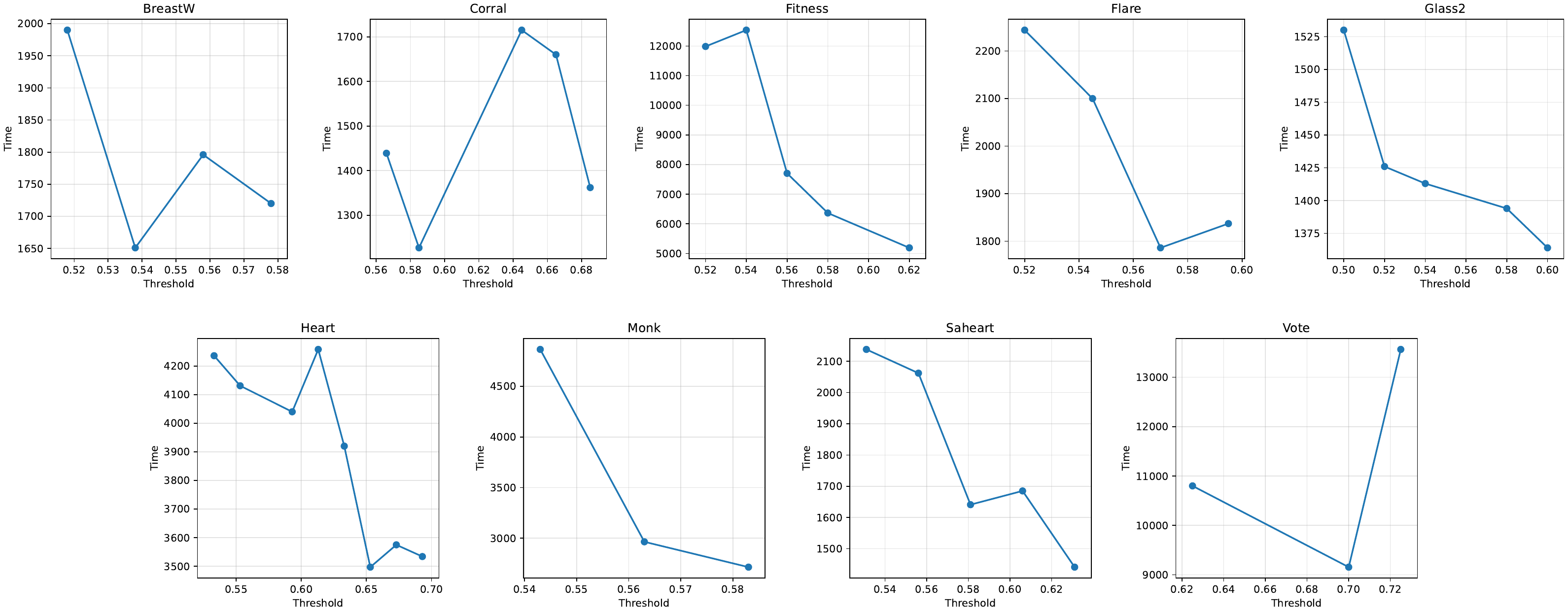}
\caption{Time obtained for the different thresholds of GSI for all datasets with noise during the training phase (PegasusQSVM).}
\label{fig:ThresVSTimeNS_QSVM}
\end{figure}

\begin{figure}[!ht]
\centering
\includegraphics[width=\textwidth]{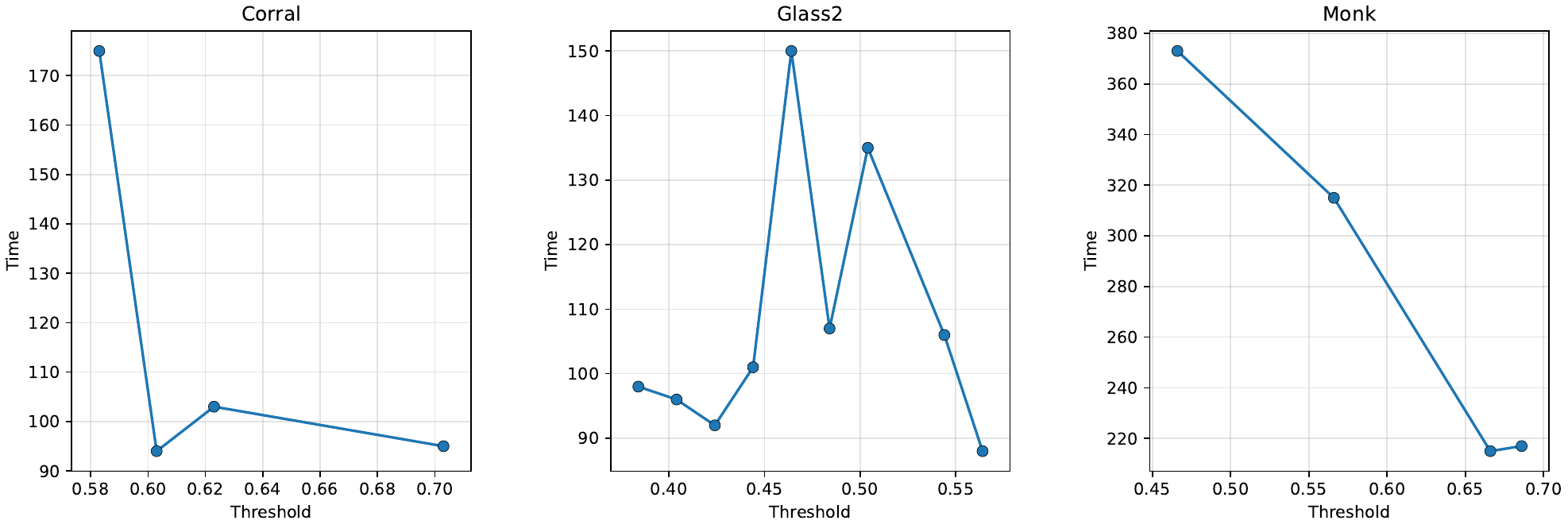}
\caption{Time obtained for the different thresholds of GSI for three datasets in a real device during the training phase (PegasusQSVM).}
\label{fig:ThresVSTimeRD_QSVM}
\end{figure}

Taken together, the three time figures (\ref{fig:ThresVSTimeFNS_QSVM}, \ref{fig:ThresVSTimeNS_QSVM}, \ref{fig:ThresVSTimeRD_QSVM}) show that computational cost tends to decrease as the threshold increases, although the trend is not equally consistent across all cases. In the ideal simulator, the decrease is generally more pronounced and gradual, and several datasets show a fairly consistent reduction in time as more restrictive thresholds are applied. In the emulator, the same general trend remains evident, but absolute times increase substantially, and trajectories appear more irregular. The results from the actual hardware follow the same general trend, even though only three datasets are available, supporting the conclusion that threshold tuning remains useful for improving efficiency beyond simulation-based environments. In all cases, the leftmost point corresponds to the baseline model, while the subsequent points represent progressively modified configurations.

At the same time, these figures make it clear that execution time does not decrease monotonically. Several datasets exhibit local oscillations or intermediate increases, indicating that time is determined not only by the threshold itself but also by the specific behavior of each dataset and the execution environment. This irregularity is more pronounced in the emulator and on actual hardware, where noise and device variability result in less stable trajectories than in the ideal simulator.

After analyzing the training results and the role of the GSI metric in model selection, the next step is to examine how this criterion modifies the structure of the quantum circuits. For illustrative purposes, this process is shown using the Glass2 dataset, with only metric A considered. It should be noted that a different fitted circuit could be obtained if the selection were based instead on criteria T or B. 

%The figures corresponding to the rest of the original and fitted quantum circuits are available in the GitHub repository referenced in the Supplementary Materials section.

\begin{figure}[!ht]
\centering
\includegraphics[width=\textwidth]{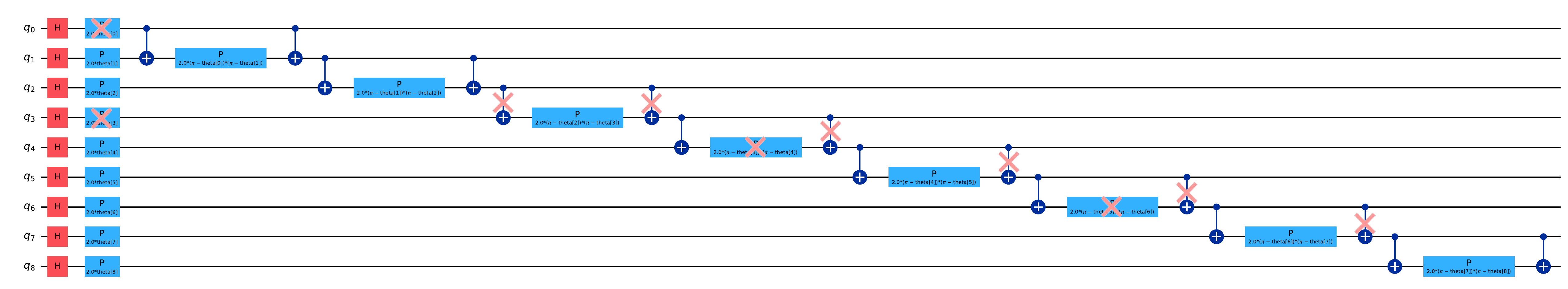}
\caption{Feature map in the original quantum circuit for Glass2 dataset.}
\label{fig:OriginalFeatureMap}
\end{figure}

Figure \ref{fig:OriginalFeatureMap} shows the original feature map of the Glass2 dataset, which contains a total of 9 qubits corresponding to its 9 input features, excluding the class label. The circuit includes several standard quantum gates, each labeled. For example, the H gate denotes the Hadamard operation, which places the qubit in a superposition; the gate represented by a filled circle connected to a plus sign enclosed in a circle corresponds to a CNOT operation, which introduces entanglement between qubits; and the P gate is used to encode classical information into quantum states. The gates marked with an \textit{X} are those selected for removal because their GSI values fall below the chosen threshold. In this case, the selected threshold is 0.560, since this configuration achieves RA=1, as indicated in Table \ref{tab:training_fns_1_dedup_QSVM}. Below this threshold, several operations are removed from the circuit, for example, the qubit $q0$ loses four gates, namely two CNOT gates, one H gate, and one P gate.

\begin{figure}[!ht]
\centering
\includegraphics[width=\textwidth]{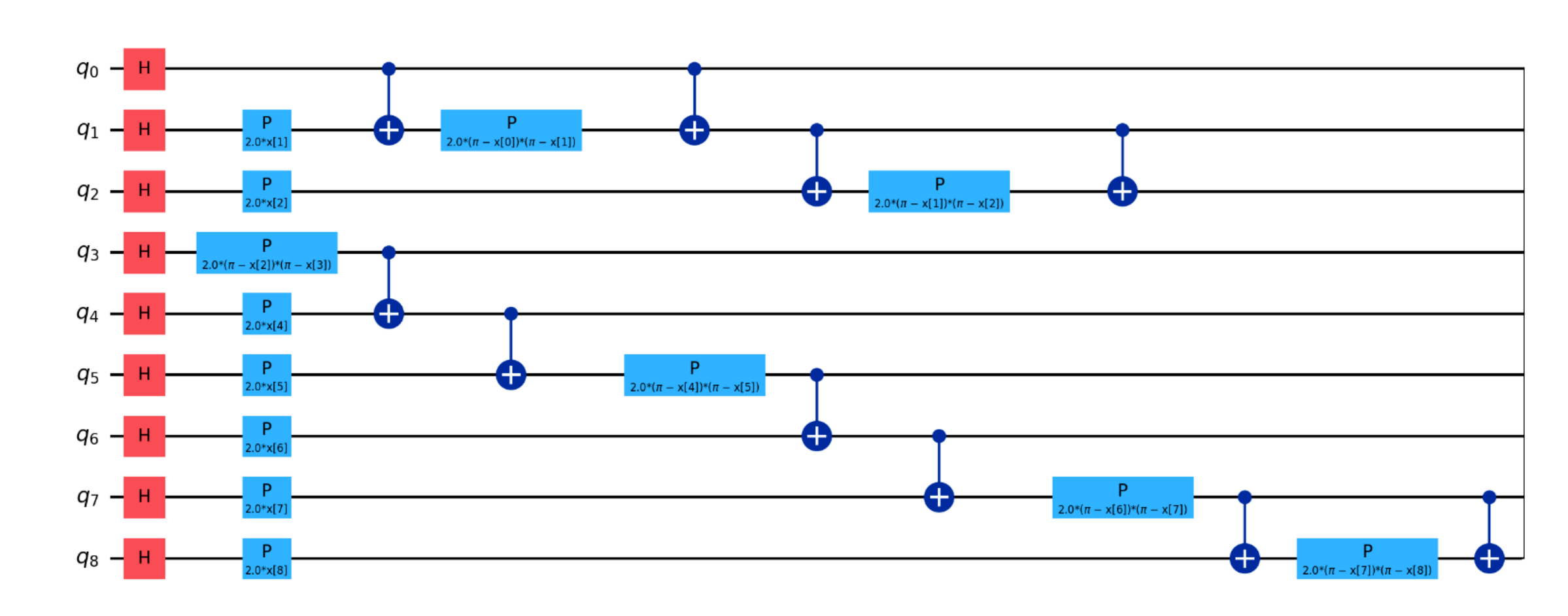}
\caption{Feature map for the most accurate quantum circuit for the Glass2 dataset.}
\label{fig:OptimizedFeatureMap}
\end{figure}

Figure \ref{fig:OptimizedFeatureMap} shows the resulting circuit after removing the selected gates and reorganizing the remaining operations. The original circuit, which contains 42 gates, is reduced to 32 gates in the optimized version, including the removal of 4 P gates and 6 CNOT gates. This reduction is particularly significant because gates such as CNOT and P contribute significantly to the circuit’s cost, both in terms of execution time and cumulative error, especially in non-ideal environments. Consequently, the optimized circuit preserves the essential transformations of the original model while achieving a more efficient and better-balanced structure.

The results obtained during the model testing phase are presented and analyzed below. As shown in Table \ref{tab:combinedTestQSVM}, the table is organized into eight columns. The first column specifies the dataset, while the second identifies the model, including the baseline model, the best model in terms of accuracy ($R_A$), the best model in terms of runtime ($R_T$), and the best model in terms of the balance between accuracy and runtime ($R_B$). The remaining columns are grouped according to the execution environment. Specifically, the third and fourth columns show the accuracy and runtime obtained in the ideal simulator (FNS), the fifth and sixth columns present the same metrics in the noisy simulator (NS), and the last two columns display the corresponding results for the real-device environment, which is only available for a limited number of datasets.

The test results show that the configurations selected during training generally retain their effectiveness when evaluated on unseen data. However, their impact is inconsistent across datasets and execution parameters. In the ideal simulator, the trained models generally outperform the baseline not only in accuracy but also in runtime, indicating that the structural changes introduced by GSI remain beneficial beyond the training phase. In many cases, the best-performing configurations are not isolated improvements in a single metric, but rather models that maintain a more favorable overall balance between predictive quality and computational cost.

When noise is introduced, the same general trend persists, but the improvements become more varied. Several datasets continue to benefit from the selected configurations, sometimes with substantial reductions in runtime and, in many cases, with equal or higher accuracy. However, the noisy environment also makes the trade-off more evident, as the configuration with the best runtime is not always the one with the highest predictive performance, and the model ranked as the best during training may not remain equally dominant during testing. This suggests that the impact of noise is not limited to reducing absolute performance but also affects the relative stability of the candidate configurations.

The results on real devices, although limited to three datasets, follow the same general trend. The baseline configuration is not consistently optimal, and the selected models can still achieve better predictive performance without causing the runtime to increase substantially. At the same time, the real device environment presents a more constrained scenario, in which differences in runtime are much smaller than in simulation-based environments, making accuracy a more decisive factor. Overall, the table confirms that the proposed selection strategy remains valid during testing, the most competitive models are typically found among the fine-tuned configurations. However, the optimal choice still depends on the specific dataset and runtime environment.

\begin{table*}[htb]
\caption{Accuracy and execution time for the original and best models during the test process with the PegasusQSVM algorithm.}
\label{tab:combinedTestQSVM}
\setlength{\tabcolsep}{4pt}
\renewcommand{\arraystretch}{1.25}
\centering
\begin{tabular}{|p{45pt}|p{55pt}|p{35pt}|p{40pt}|p{35pt}|p{40pt}|p{35pt}|p{40pt}|}
\hline
\multicolumn{2}{|c|}{} & \multicolumn{2}{c|}{\textbf{FNS}} & \multicolumn{2}{c|}{\textbf{NS}} & \multicolumn{2}{c|}{\textbf{RD}} \\
\hline
\textbf{Dataset} & \textbf{Model} & \textbf{Acc.} & \textbf{Time (s)} & \textbf{Acc.} & \textbf{Time (s)} & \textbf{Acc.} & \textbf{Time (s)} \\
\hline
\multirow{4}{*}{BreastW} 
    & Baseline      & 0.807 & 59    & 0.800 & 616   & - & - \\
    & Best $R_A$    & 0.900 & 36    & 0.885 & 512   & - & - \\
    & Best $R_T$    & 0.900 & 36    & 0.728 & 495   & - & - \\
    & Best $R_B$    & 0.900 & 36    & 0.885 & 512   & - & - \\
\hline
\multirow{4}{*}{Fitness}
    & Baseline      & 0.326 & 86    & 0.326 & 6422  & - & - \\
    & Best $R_A$    & 0.673 & 81    & 0.660 & 3271  & - & - \\
    & Best $R_T$    & 0.656 & 58    & 0.500 & 2624  & - & - \\
    & Best $R_B$    & 0.656 & 58    & 0.680 & 2971  & - & - \\
\hline
\multirow{4}{*}{Corral}
    & Baseline      & 0.562 & 1.75  & 0.937 & 108   & 0.656 & 3 \\
    & Best $R_A$    & 1.000 & 1.43  & 1.000 & 87    & 0.656 & 2 \\
    & Best $R_T$    & 1.000 & 1.29  & 1.000 & 87 & 0.718 & 2 \\
    & Best $R_B$    & 1.000 & 1.29  & 1.000 & 87 & 0.750 & 2\\
\hline
\multirow{4}{*}{Glass2}
    & Baseline      & 0.666 & 9.11  & 0.727 & 123   & 0.393 & 2 \\
    & Best $R_A$    & 0.727 & 6.73  & 0.666 & 99    & 0.515 & 2 \\
    & Best $R_T$    & 0.757 & 6.90  & 0.727 & 100 & 0.393 & 2 \\
    & Best $R_B$    & 0.757 & 6.83  & 0.727 & 101 & 0.393 &  2\\
\hline
\multirow{4}{*}{Heart}
    & Baseline      & 0.695 & 127   & 0.733 & 1541  & - & - \\
    & Best $R_A$    & 0.755 & 92    & 0.750 & 1414  & - & - \\
    & Best $R_T$    & 0.755 & 92    & 0.744 & 1402  & - & - \\
    & Best $R_B$    & 0.755 & 92    & 0.755 & 1394  & - & - \\
\hline
\multirow{4}{*}{Monk}
    & Baseline      & 0.687 & 20    & 0.714 & 1344 & 0.526 & 3 \\
    & Best $R_A$    & 0.794 & 15    & 0.785 & 708 & 0.571 & 2 \\
    & Best $R_T$    & 0.794 & 15    & 0.794 & 701   & 0.571 & 2 \\
    & Best $R_B$    & 0.607 & 15    & 0.785 & 708 & 0.553 &  2\\
\hline
\multirow{4}{*}{Flare}
    & Baseline      & 0.626 & 97    & 0.598 & 976   & - & - \\
    & Best $R_A$    & 0.841 & 75    & 0.845 & 854   & - & - \\
    & Best $R_T$    & 0.598 & 67    & 0.644 & 657   & - & -\\
    & Best $R_B$    & 0.598 & 67    & 0.654 & 653   & - & - \\
\hline
\multirow{4}{*}{Vote}
    & Baseline      & 0.724 & 103   & 0.942 & 2542  & - & - \\
    & Best $R_A$    & 0.908 & 79    & 0.908 & 2039  & - & - \\
    & Best $R_T$    & 0.919 & 80    & 0.942 & 2036  & - & - \\
    & Best $R_B$    & 0.770 & 81    & 0.942 & 2035  & - & -\\
\hline
\multirow{4}{*}{Saheart}
    & Baseline      & 0.623 & 130   & 0.503 & 485   & - & - \\
    & Best $R_A$    & 0.623 & 58    & 0.623 & 356   & - & - \\
    & Best $R_T$    & 0.623 & 58    & 0.634 & 310   & - & - \\
    & Best $R_B$    & 0.623 & 58    & 0.483 & 365   & - & - \\
\hline
\end{tabular}
\end{table*}

\begin{figure}[!ht]
\centering
\includegraphics[width=\textwidth]{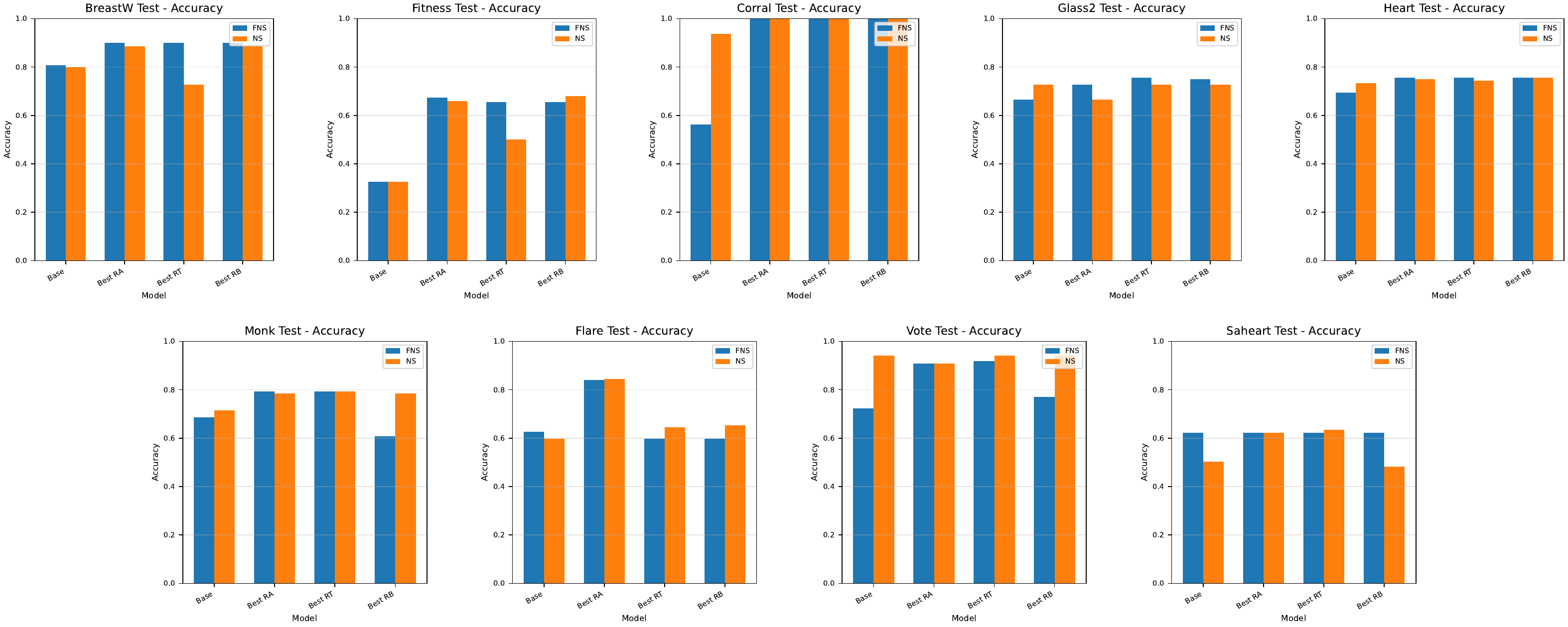}
\caption{Graph visualization of Accuracy vs. Model for different datasets with noise and without noise during the testing phase (PegasusQSVM).}
\label{fig:AccVSModel_QSVM}
\end{figure}

\begin{figure}[!ht]
\centering
\includegraphics[width=\textwidth]{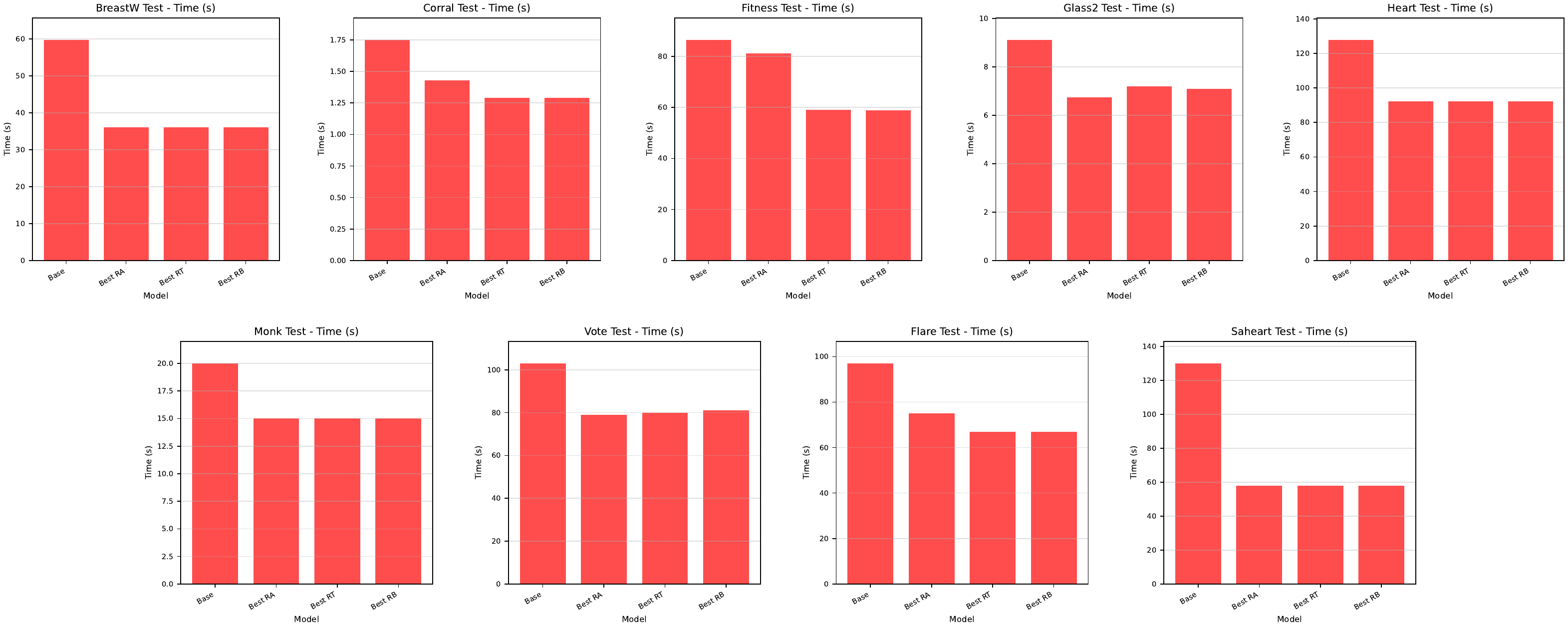}
\caption{Graph visualization of Time vs. Model for different datasets without noise during the testing phase (PegasusQSVM).}
\label{fig:TimeVSModelFNS_QSVM}
\end{figure}

\begin{figure}[!ht]
\centering
\includegraphics[width=\textwidth]{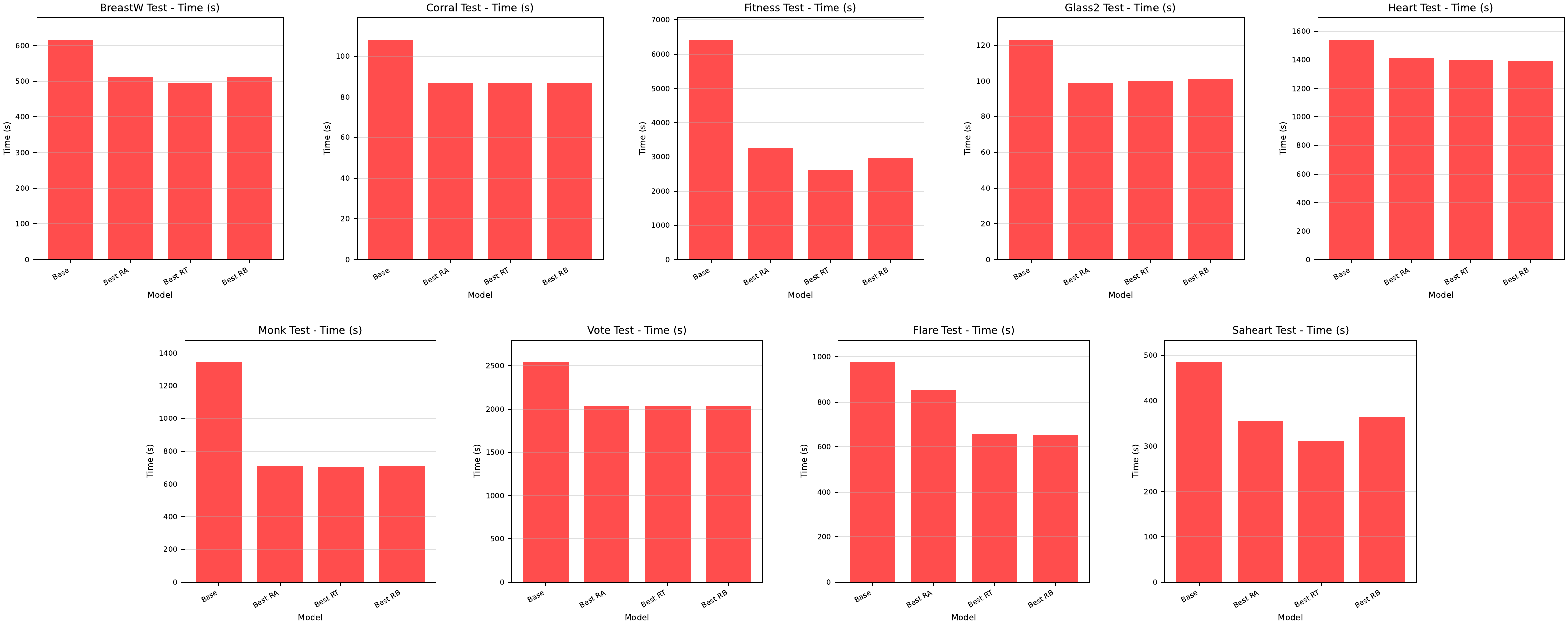}
\caption{Graph visualization of Time vs. Model for different datasets with noise during the testing phase (PegasusQSVM).}
\label{fig:TimeVSModelNS_QSVM}
\end{figure}

\begin{figure}[!ht]
\centering
\includegraphics[width=\textwidth]{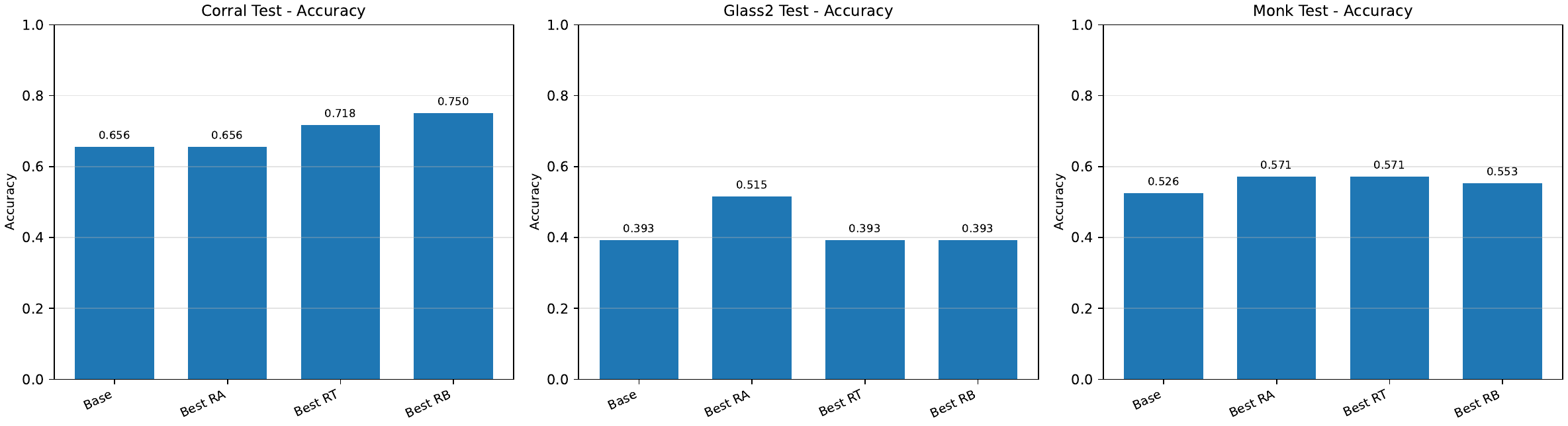}
\caption{Graph visualization of Accuracy vs. Model for three datasets in a real device during the testing phase (PegasusQSVM).}
\label{fig:AccVSModel_RD_QSVM}
\end{figure}

Figures \ref{fig:AccVSModel_QSVM}, \ref{fig:TimeVSModelFNS_QSVM}, \ref{fig:TimeVSModelNS_QSVM}, and \ref{fig:AccVSModel_RD_QSVM} provide a graphical summary of the test results presented in Table \ref{tab:combinedTestQSVM}. Overall, the selected models maintain or improve the baseline accuracy on many datasets while reducing runtime, particularly in the FNS and NS environments. The RD figure, although limited to three datasets, follows the same general trend. Taken together, these visualizations support the conclusions drawn from the table, confirming that the selected configurations generally offer a better balance between predictive performance and efficiency than the original model.

%Resultados del testeo graficados

In general, datasets associated with more demanding circuits tend to exhibit larger reductions in execution time when the selected configurations replace the baseline, whereas simpler datasets typically show smaller, though still meaningful, improvements. At the same time, the effect on accuracy is not uniform, indicating that the benefit of reducing circuit complexity depends on the extent to which each dataset relies on the transformations provided by the removed gates.

Another important factor is the sensitivity of each dataset to noise and variations in circuit structure. In some cases, the tuned circuits maintain or even improve predictive performance while reducing computational cost, suggesting that the original model contains operations that are not essential for generalization. In other cases, the benefits are more limited, or a clearer trade-off between accuracy and time is observed, especially in environments with noisy simulators and real devices. Overall, these results confirm that the proposed methodology does not produce a uniform effect across all problems, but rather adapts differently depending on the complexity, robustness, and internal structure of each dataset.

\subsubsection{Discussion of QNN results}

The simulator results for QNN (Table \ref{tab:training_fns_1_qnn} and Table \ref{tab:training_fns_2_qnn}) show that increasing the GSI value typically leads to a reduction in the number of gates and, in most cases, to shorter execution times as well. Unlike the behavior previously observed for PegasusQSVM, improvements in accuracy are present, but they are more moderate and less consistent across datasets. In several cases, the best-performing configurations are still found in intermediate or moderately reduced circuits, indicating that the original model is not always the most suitable choice and that the most compressed configuration does not consistently obtain the best overall result.

Another important observation is that the QNN results reveal a more pronounced trade-off between accuracy and efficiency. On some datasets, simplified circuits improve both metrics simultaneously. In contrast, on other datasets, the gains in runtime are accompanied only by small improvements in predictive performance, or even by a loss of accuracy. This suggests that the effect of GSI-based circuit tuning in QNNs is more sensitive to dataset-dependent behavior, and that the most favorable configurations are those that preserve a reasonable balance among circuit reduction, runtime, and predictive quality, rather than simply minimizing the number of gates.

\begin{table}[!htb]
\caption{Training results (QNN) for the first five datasets in the simulator.}
\label{tab:training_fns_1_qnn}
\setlength{\tabcolsep}{4pt}
\renewcommand{\arraystretch}{1.2}
\centering
\begin{tabular}{|l|cc|ccc|}
\hline
\textbf{Dataset} & \textbf{GSI} & \textbf{\#Gates} & \textbf{Acc.} & \textbf{Time} & \boldmath{$R_{ATB}$} \\
\hline
\multirow[c]{4}{*}{BreastW} & 0.518 & 42 & 0.578 & 105.56 & -- \\
                            & 0.538 & 33 & 0.607 & 98.27  & \textbf{1}-3-3 \\
                            & 0.558 & 29 & 0.607 & 95.26  & 2-2-\textbf{1} \\
                            & 0.578 & 28 & 0.585 & 94.27  & 3-\textbf{1}-2 \\
\hline
\multirow[c]{7}{*}{Heart}   & 0.533 & 52 & 0.543 & 351.92 & -- \\
                            & 0.553 & 48 & 0.559 & 346.78 & 6-6-6 \\
                            & 0.593 & 47 & 0.668 & 346.15 & \textbf{1}-5-\textbf{1} \\
                            & 0.613 & 45 & 0.581 & 344.07 & 3-4-5 \\
                            & 0.633 & 43 & 0.625 & 342.30 & 2-3-3 \\
                            & 0.673 & 31 & 0.576 & 329.57 & 4-2-4 \\
                            & 0.693 & 24 & 0.576 & 320.55 & 5-\textbf{1}-2 \\
\hline
\multirow[c]{2}{*}{Flare}   & 0.503 & 47 & 0.586 & 246.35 & -- \\
                            & 0.523 & 40 & 0.591 & 237.75 & \textbf{1}-\textbf{1}-\textbf{1} \\
\hline
\multirow[c]{5}{*}{Vote}    & 0.500 & 77 & 0.482 & 688.71 & -- \\
                            & 0.600 & 75 & 0.551 & 676.08 & 2-3-3 \\
                            & 0.620 & 74 & 0.551 & 672.52 & 3-2-4 \\
                            & 0.640 & 62 & 0.459 & 551.65 & 4-4-2 \\
                            & 0.660 & 59 & 0.655 & 548.70 & \textbf{1}-\textbf{1}-\textbf{1} \\
\hline
\multirow[c]{5}{*}{Glass2}  & 0.500 & 42 & 0.575 & 24.43 & -- \\
                            & 0.520 & 37 & 0.424 & 23.55 & 4-4-4 \\
                            & 0.540 & 35 & 0.636 & 23.09 & \textbf{1}-3-2 \\
                            & 0.580 & 32 & 0.575 & 22.42 & 3-2-3 \\
                            & 0.600 & 30 & 0.636 & 22.38 & 2-\textbf{1}-\textbf{1} \\
\hline
\end{tabular}
\end{table}

\begin{table}[!htb]
\caption{Training results (QNN) for the last four datasets in the simulator.}
\label{tab:training_fns_2_qnn}
\setlength{\tabcolsep}{4pt}
\renewcommand{\arraystretch}{1.2}
\centering
\begin{tabular}{|l|cc|ccc|}
\hline
\textbf{Dataset} & \textbf{GSI} & \textbf{\#Gates} & \textbf{Acc.} & \textbf{Time} & \boldmath{$R_{ATB}$} \\
\hline
\multirow[c]{5}{*}{Fitness}  & 0.520 & 27 & 0.480 & 100.51 & -- \\
                             & 0.540 & 26 & 0.533 & 98.24  & 2-4-3 \\
                             & 0.560 & 21 & 0.440 & 90.53  & 4-3-4 \\
                             & 0.580 & 18 & 0.513 & 86.07  & 3-2-2 \\
                             & 0.620 & 16 & 0.550 & 83.26  & \textbf{1}-\textbf{1}-\textbf{1} \\
\hline
\multirow[c]{3}{*}{Monk}     & 0.543 & 26 & 0.531 & 36    & -- \\
                             & 0.563 & 23 & 0.729 & 34    & \textbf{1}-2-\textbf{1} \\
                             & 0.583 & 18 & 0.567 & 31    & 2-\textbf{1}-2 \\
\hline
\multirow[c]{5}{*}{Saheart}  & 0.531 & 41 & 0.537 & 69.97 & -- \\
                             & 0.551 & 35 & 0.537 & 66.63 & 2-4-4 \\
                             & 0.571 & 30 & 0.559 & 63.63 & \textbf{1}-3-\textbf{1} \\
                             & 0.591 & 29 & 0.526 & 63.31 & 3-2-3 \\
                             & 0.611 & 26 & 0.516 & 61.75 & 4-\textbf{1}-2 \\
\hline
\multirow[c]{5}{*}{Corral}   & 0.565 & 27 & 0.280 & 11.00 & -- \\
                             & 0.605 & 24 & 0.750 & 10.73 & \textbf{1}-4-2 \\
                             & 0.645 & 20 & 0.750 & 10.08 & 2-3-\textbf{1} \\
                             & 0.665 & 19 & 0.625 & 10.05 & 3-2-3 \\
                             & 0.685 & 14 & 0.531 & 9.46  & 4-\textbf{1}-4 \\
\hline
\end{tabular}
\end{table}

The results from the QNN emulator (Table \ref{tab:training_ns_1_qnn} and Table \ref{tab:training_ns_2_qnn}) confirm that the GSI effect remains useful in the presence of noise. However, the behavior is clearly less consistent than in the ideal simulator. In several cases, the best-performing configurations are again found in intermediate or moderately reduced circuits, demonstrating that the reference configuration is not systematically the most favorable option. At the same time, the noisy environment makes the differences between candidate configurations more pronounced, especially in datasets where accuracy fluctuates more as circuit complexity decreases.

A key finding from these results is that noise constrains the trade-off between predictive quality and efficiency. On some datasets, the selected configurations continue to improve both accuracy and runtime, whereas on others, the reduction in runtime is greater than the gain in predictive performance. This suggests that, for QNNs in the emulator, GSI is particularly valuable for controlling circuit cost while maintaining competitive performance. However, the ultimate benefit ultimately depends largely on each dataset’s sensitivity to noise and structural changes in the circuit.

\begin{table}[!htb]
\caption{Training results (QNN) for the first five datasets in the emulator.}
\label{tab:training_ns_1_qnn}
\setlength{\tabcolsep}{4pt}
\renewcommand{\arraystretch}{1.2}
\centering
\begin{tabular}{|l|cc|ccc|}
\hline
\textbf{Dataset} & \textbf{GSI} & \textbf{\#Gates} & \textbf{Acc.} & \textbf{Time} & \boldmath{$R_{ATB}$} \\
\hline
\multirow[c]{5}{*}{BreastW} & 0.430 & 42 & 0.521 & 1330 & -- \\
                            & 0.450 & 41 & 0.557 & 1329 &  2-4-4 \\
                            & 0.510 & 35 & 0.535 & 1191 &  4-\textbf{1}-2 \\
                            & 0.530 & 31 & 0.657 & 1191 &  \textbf{1}-2-\textbf{1} \\
                            & 0.550 & 29 & 0.542 & 1272 &  3-3-3 \\
\hline
\multirow[c]{2}{*}{Heart}   & 0.378 & 52 & 0.527 & 4042 & -- \\
                            & 0.518 & 49 & 0.586 & 4026 & \textbf{1}-\textbf{1}-\textbf{1} \\
\hline
\multirow[c]{4}{*}{Flare}   & 0.360 & 47 & 0.455 & 2244 & -- \\
                            & 0.380 & 42 & 0.497 & 1953 & 2-3-3 \\
                            & 0.400 & 35 & 0.488 & 1849 & 3-2-2 \\
                            & 0.440 & 32 & 0.516 & 1834 & \textbf{1}-\textbf{1}-\textbf{1} \\
\hline
\multirow[c]{4}{*}{Vote}    & 0.552 & 77 & 0.609 & 2943 & -- \\
                            & 0.620 & 74 & 0.448 & 2755 & 2-3-3 \\
                            & 0.640 & 62 & 0.459 & 2239 & \textbf{1}-\textbf{1}-\textbf{1} \\
                            & 0.660 & 59 & 0.448 & 2251 & 3-2-2 \\
\hline
\multirow[c]{7}{*}{Glass2}  & 0.334 & 42 & 0.303 & 385 & -- \\
                            & 0.394 & 37 & 0.545 & 350 & 2-4-2 \\
                            & 0.434 & 35 & 0.363 & 350 & 5-3-4 \\
                            & 0.454 & 32 & 0.515 & 368 & 4-6-3 \\
                            & 0.494 & 31 & 0.363 & 366 & 6-5-5 \\
                            & 0.514 & 25 & 0.545 & 348 & 3-2-\textbf{1} \\
                            & 0.534 & 25 & 0.575 & 346 & \textbf{1}-\textbf{1}-6 \\
\hline
\end{tabular}
\end{table}

\begin{table}[!htb]
\caption{Training results (QNN) for the last four datasets in the emulator.}
\label{tab:training_ns_2_qnn}
\setlength{\tabcolsep}{4pt}
\renewcommand{\arraystretch}{1.2}
\centering
\begin{tabular}{|l|cc|ccc|}
\hline
\textbf{Dataset} & \textbf{GSI} & \textbf{\#Gates} & \textbf{Acc.} & \textbf{Time} & \boldmath{$R_{ATB}$} \\
\hline
\multirow[c]{5}{*}{Fitness}  & 0.360 & 27 & 0.513 & 16918 & -- \\
                             & 0.500 & 23 & 0.646 & 16620 & \textbf{1}-3-4 \\
                             & 0.680 & 18 & 0.536 & 12590 & 2-2-\textbf{1} \\
                             & 0.700 & 15 & 0.503 & 12422 & 3-\textbf{1}-2 \\
                             & 0.740 & 14 & 0.466 & 11948 & 4-4-3 \\
\hline
\multirow[c]{6}{*}{Monk}     & 0.391 & 26 & 0.459 & 5163 & -- \\
                             & 0.431 & 25 & 0.378 & 5008 & 5-4-5 \\
                             & 0.511 & 23 & 0.549 & 5080 & 2-5-4 \\
                             & 0.671 & 19 & 0.540 & 3324 & 3-2-2 \\
                             & 0.691 & 17 & 0.594 & 3314 & \textbf{1}-\textbf{1}-\textbf{1} \\
                             & 0.711 & 16 & 0.495 & 3712 & 4-3-3 \\
\hline
\multirow[c]{9}{*}{Saheart}  & 0.405 & 42 & 0.526 & 915 & -- \\
                             & 0.425 & 41 & 0.526 & 914 & 4-8-8 \\
                             & 0.505 & 38 & 0.548 & 913 & 2-7-5 \\
                             & 0.665 & 37 & 0.537 & 906 & 3-6-6 \\
                             & 0.685 & 32 & 0.494 & 879 & 6-5-7 \\
                             & 0.705 & 29 & 0.483 & 809 & 7-3-4 \\
                             & 0.725 & 26 & 0.505 & 815 & 5-4-3 \\
                             & 0.765 & 24 & 0.580 & 799 & \textbf{1}-2-\textbf{1} \\
                             & 0.785 & 22 & 0.483 & 745 & 8-\textbf{1}-2 \\
\hline
\multirow[c]{3}{*}{Corral}   & 0.430 & 27 & 0.406 & 1556 & -- \\
                             & 0.450 & 26 & 0.500 & 1555 & 2-2-2 \\
                             & 0.510 & 24 & 0.562 & 1526 & \textbf{1}-\textbf{1}-\textbf{1} \\
\hline
\end{tabular}
\end{table}

The results obtained on real devices (Table \ref{tab:training_RD_qnn}) for QNNs show more limited performance compared to simulator and emulator environments, which is to be expected under hardware execution conditions. In this case, the reductions in the number of logic gates are smaller and the differences in execution time are also more limited. Therefore, the effect of GSI is more clearly manifested in precision gains than in large efficiency gains. Nevertheless, the baseline configuration is not always optimal, as some tuned circuits continue to achieve better overall results, particularly on datasets such as Glass2 and Monk.

At the same time, these results indicate that the advantage of GSI under real-device conditions is more pronounced and depends heavily on the dataset. In some cases, moderate adjustments remain useful and lead to more balanced models, while in others the baseline model remains competitive or the improvements are only marginal. Therefore, the real device environment confirms the same general idea observed in previous environments, but in a more restrictive scenario. Since GSI continues to help identify more suitable circuit configurations, although its effect becomes less consistent and more sensitive to hardware-level variability.

\begin{table}[!htb]
\caption{Training results (QNN) for three datasets in a real device.}
\label{tab:training_RD_qnn}
\setlength{\tabcolsep}{4pt}
\renewcommand{\arraystretch}{1.2}
\centering
\begin{tabular}{|l|cc|ccc|}
\hline
\textbf{Dataset} & \textbf{GSI} & \textbf{\#Gates} & \textbf{Acc.} & \textbf{Time} & \boldmath{$R_{ATB}$} \\
\hline
\multirow[c]{4}{*}{Corral}   & 0.500 & 27 & 0.656 & 214 & -- \\
                             & 0.520 & 26 & 0.593 & 200 & \textbf{1}-\textbf{1}-\textbf{1} \\
                             & 0.580 & 24 & 0.437 & 238 & 2-2-3 \\
                             & 0.660 & 23 & 0.375 & 204 & 3-3-2 \\
\hline
\multirow[c]{6}{*}{Glass2}   & 0.366 & 42 & 0.545 & 210 & -- \\
                             & 0.406 & 40 & 0.606 & 208 & \textbf{1}-5-\textbf{1} \\
                             & 0.426 & 39 & 0.484 & 201 & 3-2-4 \\
                             & 0.486 & 38 & 0.464 & 202 & 5-4-5 \\
                             & 0.546 & 37 & 0.515 & 201 & 2-3-2 \\
                             & 0.566 & 35 & 0.484 & 200 & 4-\textbf{1}-3 \\
\hline
\multirow[c]{3}{*}{Monk}     & 0.433 & 27 & 0.495 & 317 & -- \\
                             & 0.453 & 26 & 0.459 & 316 & 2-2-2 \\
                             & 0.573 & 25 & 0.585 & 246 & \textbf{1}-\textbf{1}-\textbf{1}  \\
                             
\hline
\end{tabular}
\end{table}

\begin{figure}[!ht]
\centering
\includegraphics[width=\textwidth]{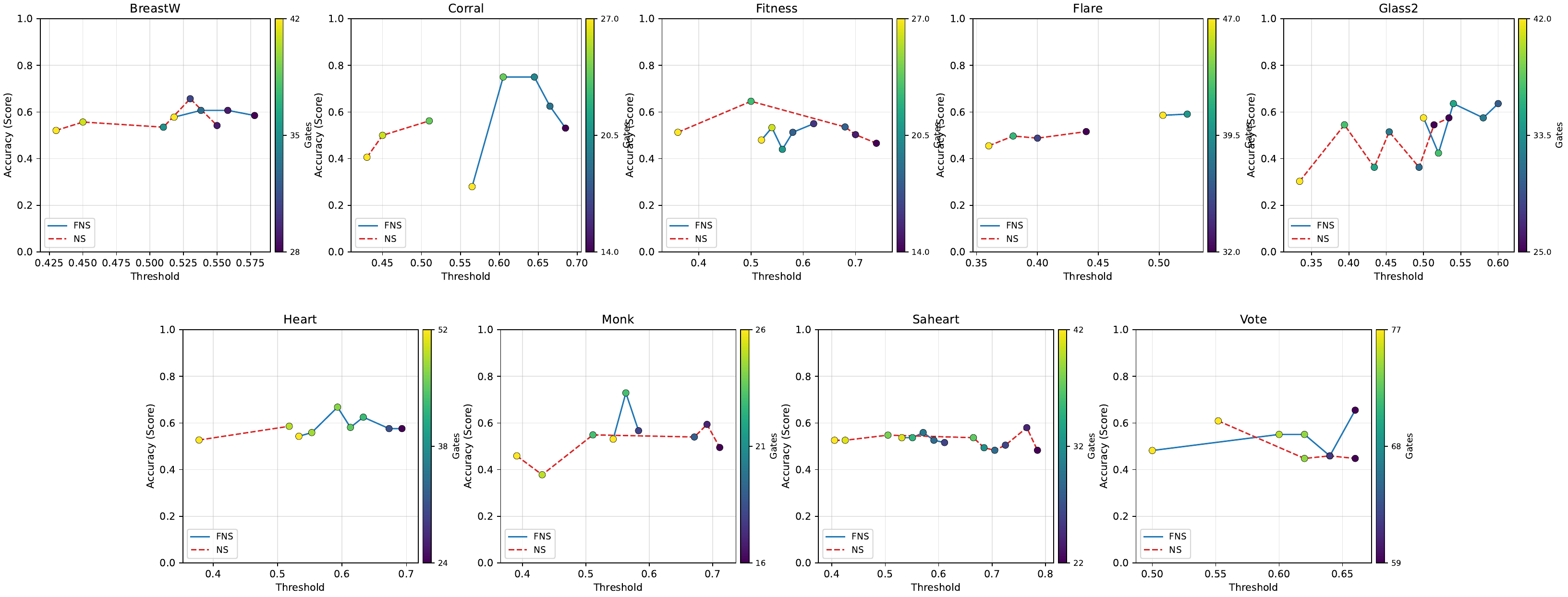}
\caption{Accuracy obtained for the different thresholds of GSI for all datasets with noise and without noise during the training phase (QNN).}
\label{fig:ThresVSAcc_QNN}
\end{figure}

\begin{figure}[!ht]
\centering
\includegraphics[width=\textwidth]{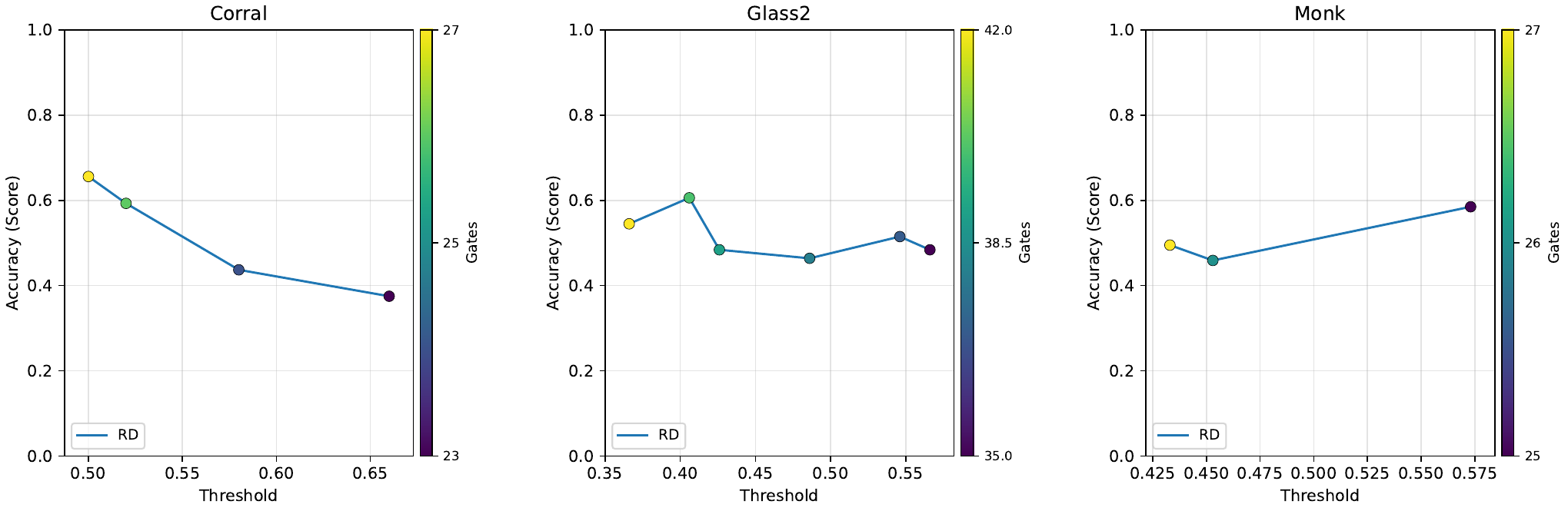}
\caption{Accuracy obtained for the different thresholds of GSI for three datasets in a real device during the training phase (QNN).}
\label{fig:ThresVSAccRD_QNN}
\end{figure}

The accuracy trends shown in Figure \ref{fig:ThresVSAcc_QNN} are consistent with the patterns already observed in the training tables. In simulator and emulator environments, the best QNN results are generally obtained with intermediate threshold values rather than with the baseline configuration or the most reduced configurations, suggesting that the most suitable models typically maintain a balanced level of circuit complexity. The results on real devices in Figure \ref{fig:ThresVSAccRD_QNN} follow the same general trend, albeit with more irregular trajectories, which is consistent with the greater variability of hardware execution.

\begin{figure}[!ht]
\centering
\includegraphics[width=\textwidth]{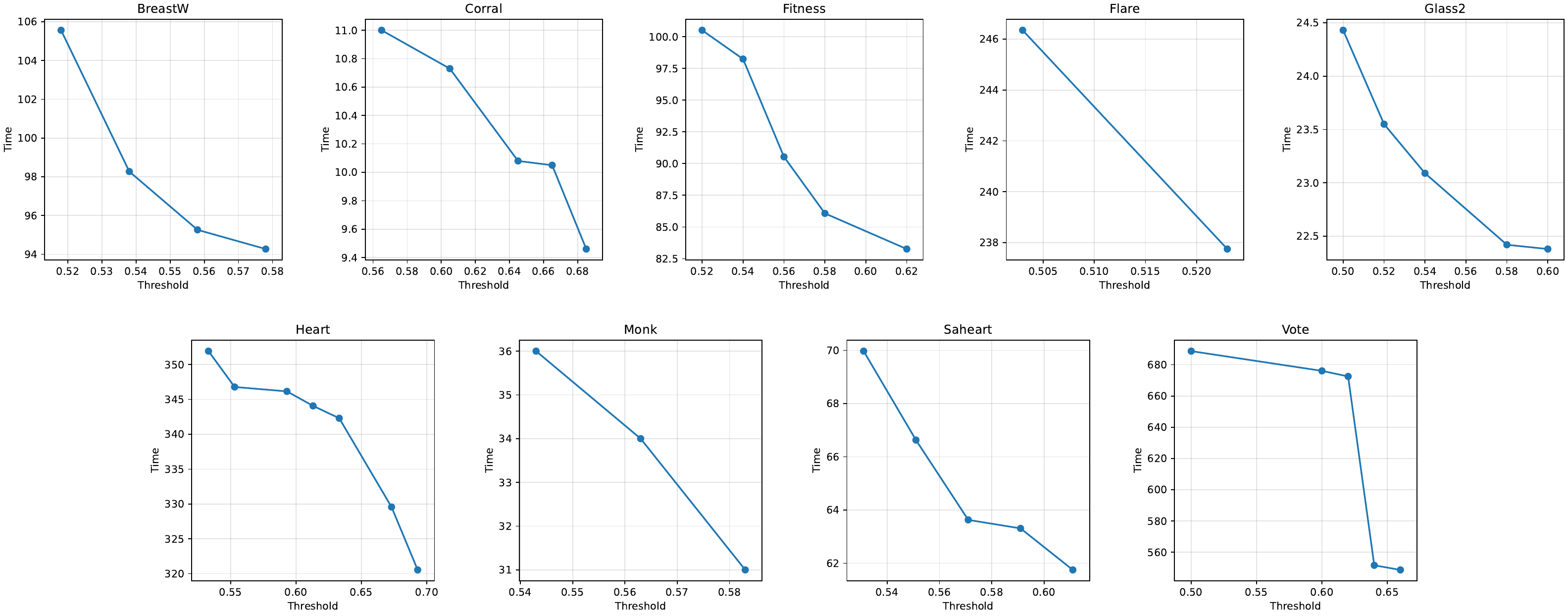}
\caption{Time obtained for the different thresholds of GSI for all datasets without noise during the training phase (QNN).}
\label{fig:ThresVSTimeFNS_QNN}
\end{figure}

\begin{figure}[!ht]
\centering
\includegraphics[width=\textwidth]{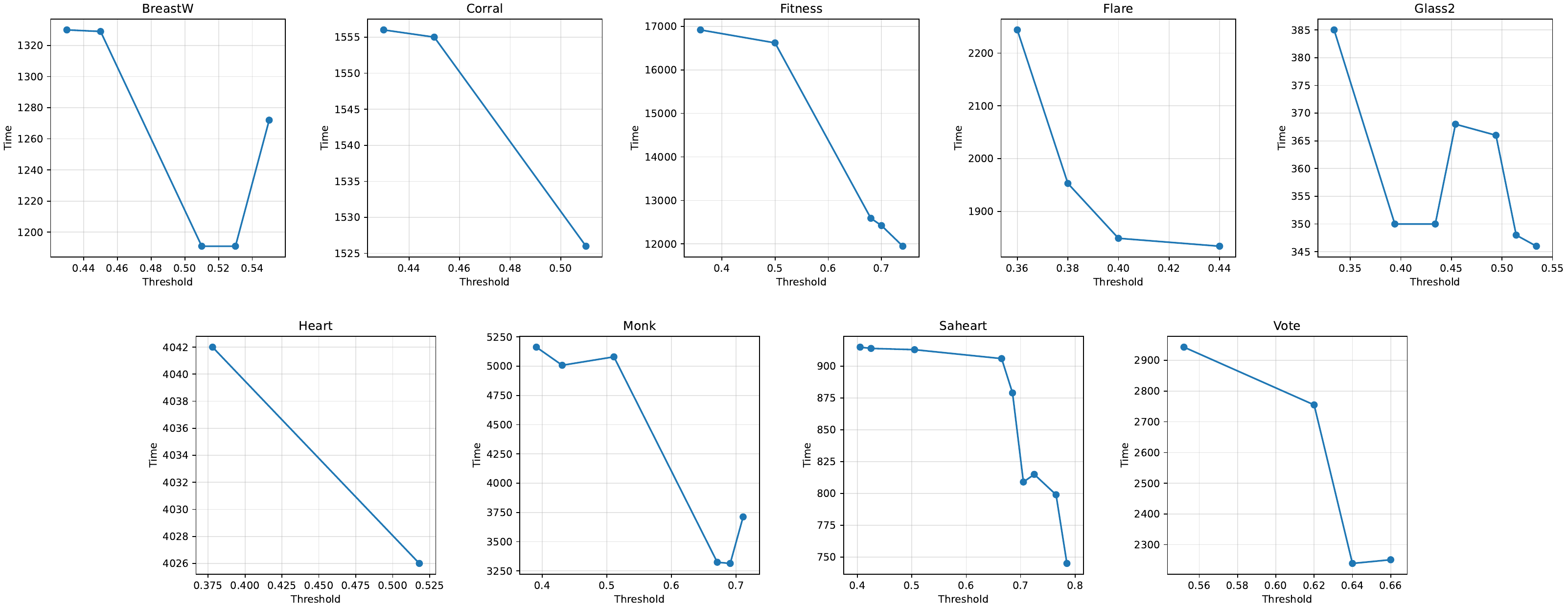}
\caption{Time obtained for the different thresholds of GSI for all datasets with noise during the training phase (QNN).}
\label{fig:ThresVSTimeNS_QNN}
\end{figure}

A similar pattern is observed in the time-domain results shown in Figures \ref{fig:ThresVSTimeFNS_QNN} and \ref{fig:ThresVSTimeNS_QNN}, as well as in the time-domain plot for the real device. In general, increasing the threshold tends to reduce runtime, with smoother behavior in the ideal simulator and a more irregular evolution in noisy and real-device environments. Taken together, these figures reinforce the interpretation from the tables: for QNNs, GSI is useful not only for reducing circuit cost but also for identifying threshold values that provide a more appropriate balance between predictive performance and efficiency.

\begin{figure}[!ht]
\centering
\includegraphics[width=\textwidth]{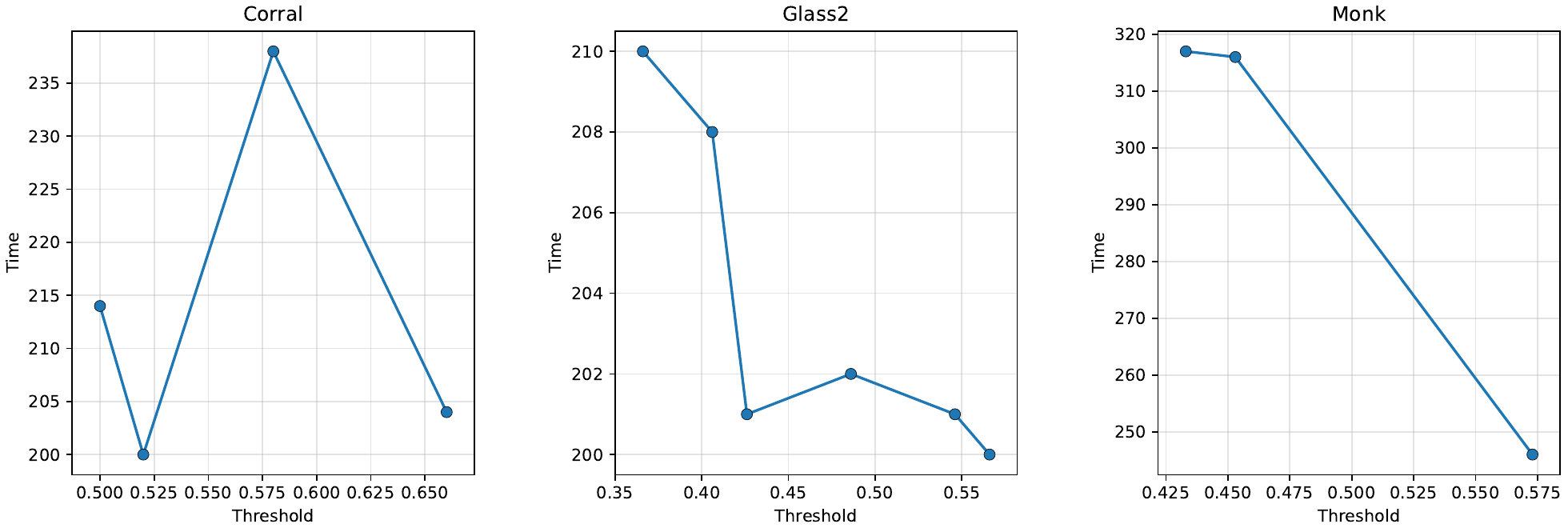}
\caption{Time obtained for the different thresholds of GSI for three datasets in real device during the training phase (QNN).}
\label{fig:ThresVSTimeRD_QNN}
\end{figure}

The results of the QNN tests (Table \ref{tab:combinedTest}) show that the configurations selected during training generally remain competitive on unseen data. However, the relationship between training and evaluation is not always straightforward. In both the FNS and NS configurations, the trained models generally reduce runtime relative to the baseline, whereas improvements in accuracy depend on the dataset. In several cases, the trend observed during training persists, and the selected configurations continue to offer a better balance between predictive performance and efficiency. However, the table also shows that the classification achieved in training does not always transfer exactly in the same way to the test phase, suggesting that the effect of GSI is more pronounced in QNN than in PegasusQSVM.

This is most clearly evident in the RD setting. For example, in Corral’s training results, the baseline remained more robust than the tuned alternatives. In contrast, in the testing phase, the baseline becomes less favorable and the selected models achieve better predictive performance. A similar shift is observed in Glass2, where the tuned configurations perform more competitively during the test phase than the original model. Therefore, the test results confirm the usefulness of the proposed selection strategy and indicate that, for QNNs, the final performance of each configuration depends greatly on how well the training trends generalize to unseen data.

\begin{table*}[htb]
\caption{Accuracy and Execution Time for the original and best models during the test process.}
\label{tab:combinedTest}
\setlength{\tabcolsep}{6pt}
\renewcommand{\arraystretch}{1.25}
\centering
\begin{tabular}{|c|c|c|c|c|c|c|c|}
\hline
\multicolumn{2}{|c|}{} & \multicolumn{2}{c|}{\textbf{FNS}} & \multicolumn{2}{c|}{\textbf{NS}} & \multicolumn{2}{c|}{\textbf{RD}} \\
\hline
\textbf{Dataset} & \textbf{Model} & \textbf{Acc.} & \textbf{Time (s)} & \textbf{Acc.} & \textbf{Time (s)} & \textbf{Acc.} & \textbf{Time (s)} \\
\hline
\multirow{4}{*}{BreastW} 
    & Baseline      & 0.578 & 0.40 & 0.507 & 10.50 & - & - \\
    & Best $R_A$    & 0.592 & 0.32 & 0.671 & 9.92 & - & - \\
    & Best $R_T$    & 0.650 & 0.30 & 0.557 & 8.50 & - & - \\
    & Best $R_B$    & 0.671 & 0.36 & 0.650 & 10.55 & - & - \\
\hline
\multirow{4}{*}{Fitness}
    & Baseline      & 0.456 & 0.42 & 0.490 & 114.00 & - & - \\
    & Best $R_A$    & 0.586 & 0.27 & 0.583 & 111.00 & - & - \\
    & Best $R_T$    & 0.586 & 0.27 & 0.470 & 81.00 & - & - \\
    & Best $R_B$    & 0.586 & 0.27 & 0.460 & 83.00 & - & - \\
\hline
\multirow{4}{*}{Corral}
    & Baseline      & 0.500 & 0.04 & 0.500 & 11.66  & 0.500 & 2\\
    & Best $R_A$    & 0.750 & 0.03 & 0.687 & 11.74 & 0.562 & 2\\
    & Best $R_T$    & 0.625 & 0.03 & 0.687 & 11.23 & 0.531 & 2\\
    & Best $R_B$    & 0.625 & 0.03 & 0.687 & 11.19 & 0.625 & 2\\
\hline
\multirow{4}{*}{Glass2}
    & Baseline      & 0.454 & 0.08 & 0.303 & 4.83 & 0.575 & 2 \\
    & Best $R_A$    & 0.666 & 0.08 & 0.484 & 4.03 & 0.666 & 2 \\
    & Best $R_T$    & 0.484 & 0.07 & 0.424 & 3.97 & 0.454 & 2 \\
    & Best $R_B$    & 0.484 & 0.07 & 0.454 & 2.89 & 0.656 & 2  \\
\hline
\multirow{4}{*}{Heart}
    & Baseline      & 0.505 & 1.20 & 0.500 & 28.00 & - & -\\
    & Best $R_A$    & 0.652 & 1.15 & 0.521 & 28.00 & - & - \\
    & Best $R_T$    & 0.527 & 1.06 & 0.521 & 28.00 & - & - \\
    & Best $R_B$    & 0.652 & 1.13 & 0.532 & 28.00 & - & - \\
\hline
\multirow{4}{*}{Monk}
    & Baseline      & 0.589 & 0.11 & 0.500 & 45.00 & 0.410 & 2 \\
    & Best $R_A$    & 0.705 & 0.16 & 0.544 & 32.00 & 0.517 & 2 \\
    & Best $R_T$    & 0.580 & 0.10 & 0.553 & 32.00 & 0.501 & 2 \\
    & Best $R_B$    & 0.580 & 0.10 & 0.616 & 32.00 & 0.526 & 2 \\
\hline
\multirow{4}{*}{Flare}
    & Baseline      & 0.602 & 0.81 & 0.518 & 17.18 & - & - \\
    & Best $R_A$    & 0.626 & 0.79 & 0.462 & 13.55 & - & - \\
    & Best $R_T$    & 0.626 & 0.79 & 0.471 & 13.53 & - & - \\
    & Best $R_B$    & 0.626 & 0.79 & 0.510 & 13.51 & - & - \\
\hline
\multirow{4}{*}{Vote}
    & Baseline      & 0.494 & 2.25 & 0.482 & 22.81 & - & - \\
    & Best $R_A$    & 0.517 & 1.81 & 0.540 & 17.55 & - & - \\
    & Best $R_T$    & 0.517 & 1.82 & 0.540 & 17.55 & - & - \\
    & Best $R_B$    & 0.517 & 1.81 & 0.540 & 18.23 & - & - \\
\hline
\multirow{4}{*}{Saheart}
    & Baseline      & 0.623 & 0.22 & 0.494 & 8.65 & - & - \\
    & Best $R_A$    & 0.397 & 0.20 & 0.537 & 6.03 & - & - \\
    & Best $R_T$    & 0.397 & 0.19 & 0.548 & 7.00 & - & - \\
    & Best $R_B$    & 0.397 & 0.19 & 0.537 & 6.01 & - & - \\
\hline
\end{tabular}
\end{table*}

\begin{figure}[!ht]
\centering
\includegraphics[width=\textwidth]{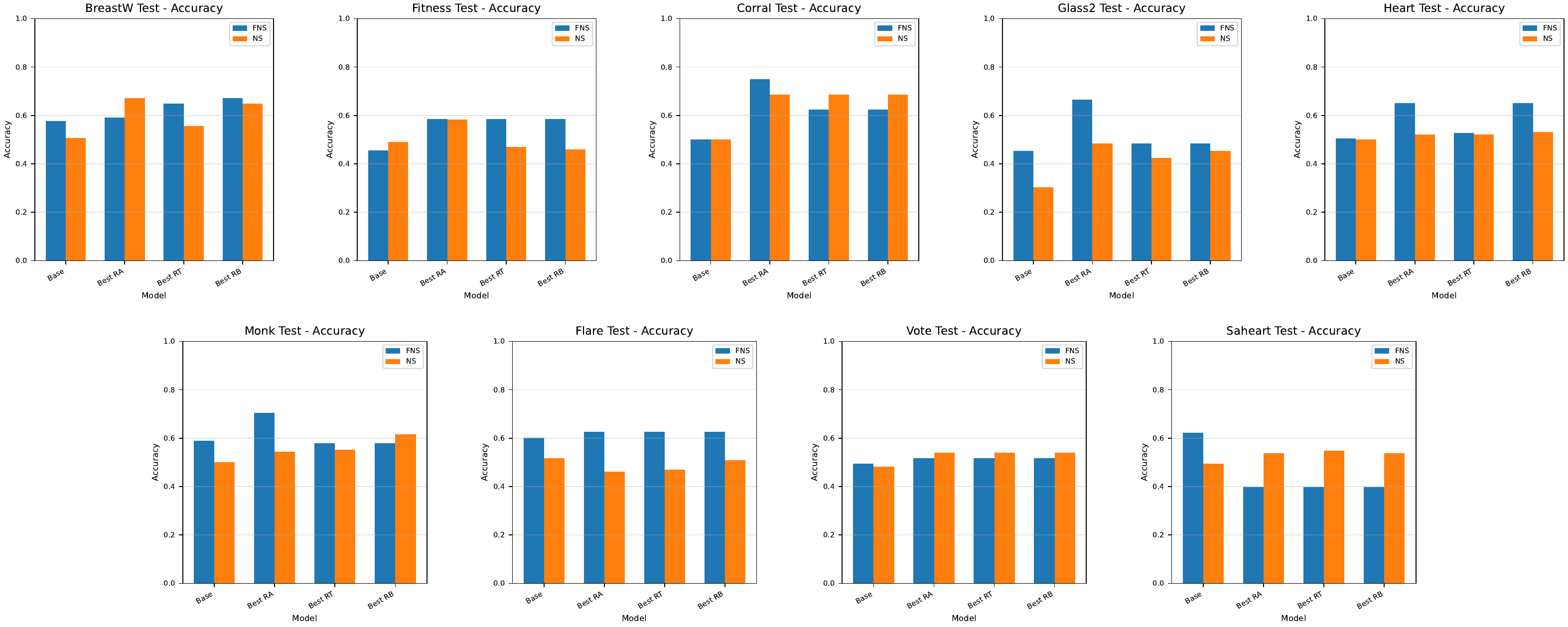}
\caption{Graph visualization of Accuracy vs. Model for different datasets with noise and without noise during the testing phase (QNN).}
\label{fig:AccVSModel_QNN}
\end{figure}

\begin{figure}[!ht]
\centering
\includegraphics[width=\textwidth]{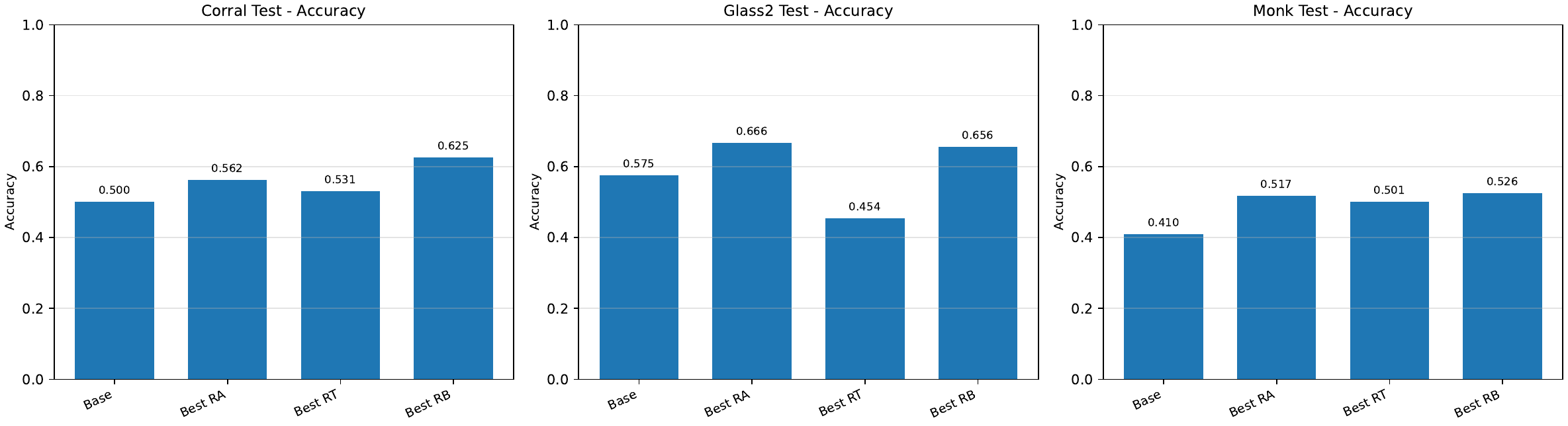}
\caption{Graph visualization of Accuracy vs. Model for three datasets in a real device during the testing phase (QNN).}
\label{fig:AccVSModel_RD_QNN}
\end{figure}

Figures \ref{fig:AccVSModel_QNN} and \ref{fig:AccVSModel_RD_QNN} visually confirm the trends already observed in Table \ref{tab:combinedTest}. In the FNS and NS environments, the selected models are generally competitive with or outperform the baseline model in terms of accuracy, although performance clearly depends on the dataset. The RD figure follows the same general pattern, showing that the baseline is not always the best choice and that, for datasets such as Corral, Glass2, and Monk, some selected configurations achieve better predictive performance than the original model. 

\begin{figure}[!ht]
\centering
\includegraphics[width=\textwidth]{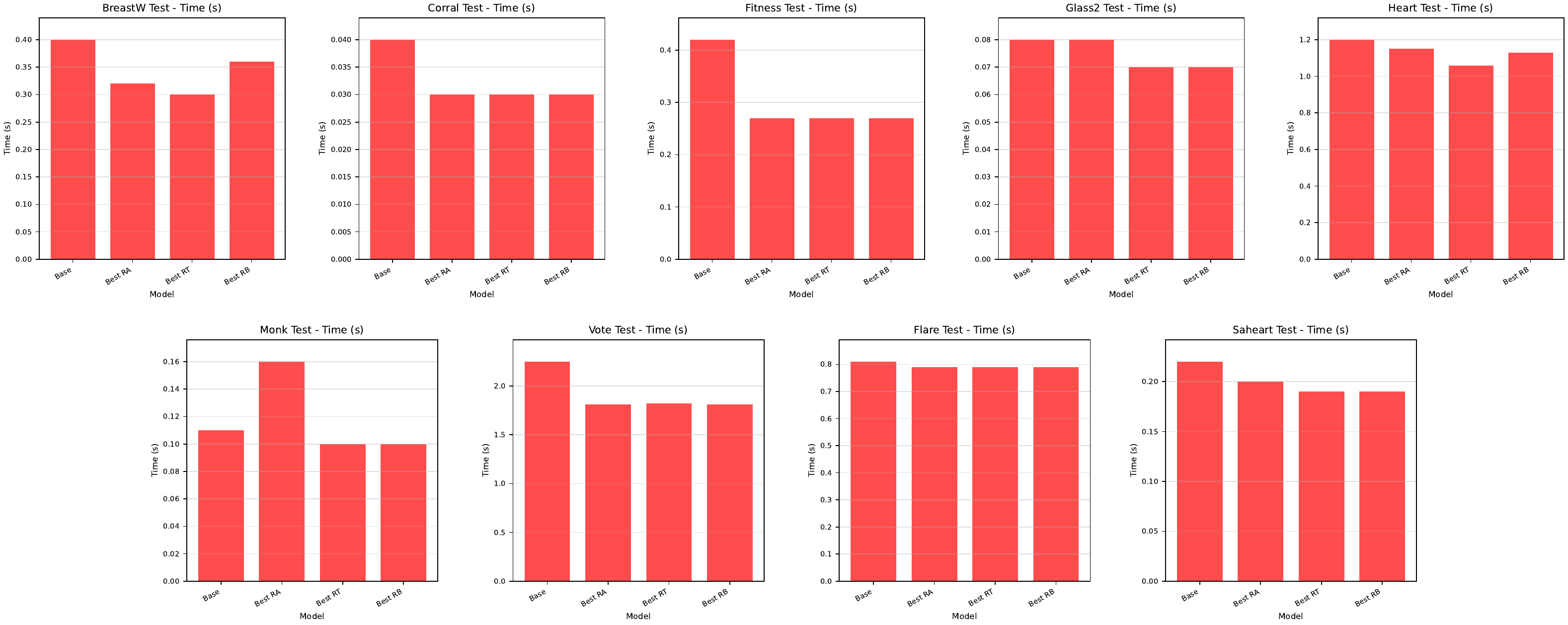}
\caption{Graph visualization of Time vs. Model for different datasets without noise during the testing phase (QNN).}
\label{fig:TimeVSModelFNS_QNN}
\end{figure}

\begin{figure}[!ht]
\centering
\includegraphics[width=\textwidth]{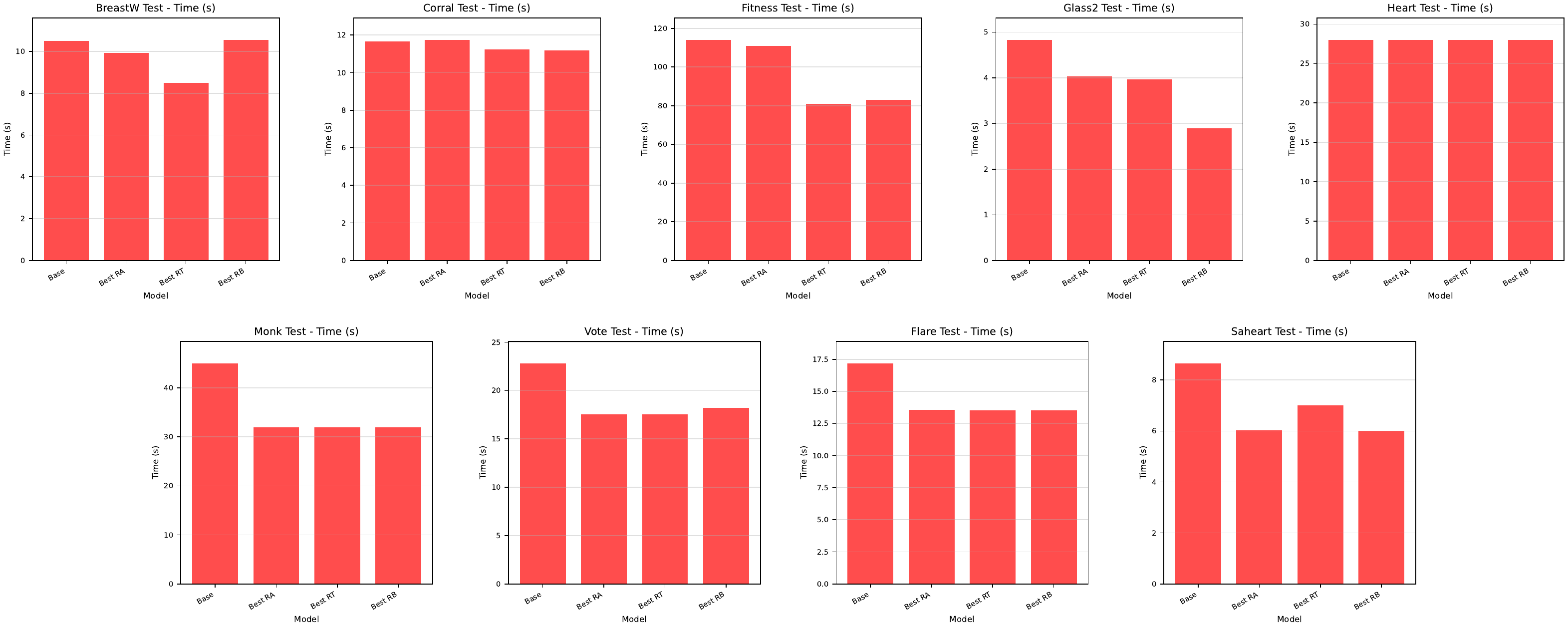}
\caption{Graph visualization of Time vs. Model for different datasets with noise during the testing phase (QNN).}
\label{fig:TimeVSModelNS_QNN}
\end{figure}

A similar conclusion can be drawn from Figures \ref{fig:TimeVSModelFNS_QNN} and \ref{fig:TimeVSModelNS_QNN}. In general, the selected models reduce runtime compared to the baseline, with smaller differences in the FNS configuration and higher absolute costs in the NS configuration. Taken together, these visualizations reinforce the interpretation of the test table. For QNN, the selected configurations generally offer a more favorable balance between predictive performance and efficiency than the baseline, although the final benefit still depends on the dataset and the execution environment.

%Resultados del testeo graficados

\subsection{Scalability analysis}\label{sec:scalabilityResults}

\begin{figure}[!ht]
\centering
\includegraphics[width=\textwidth]{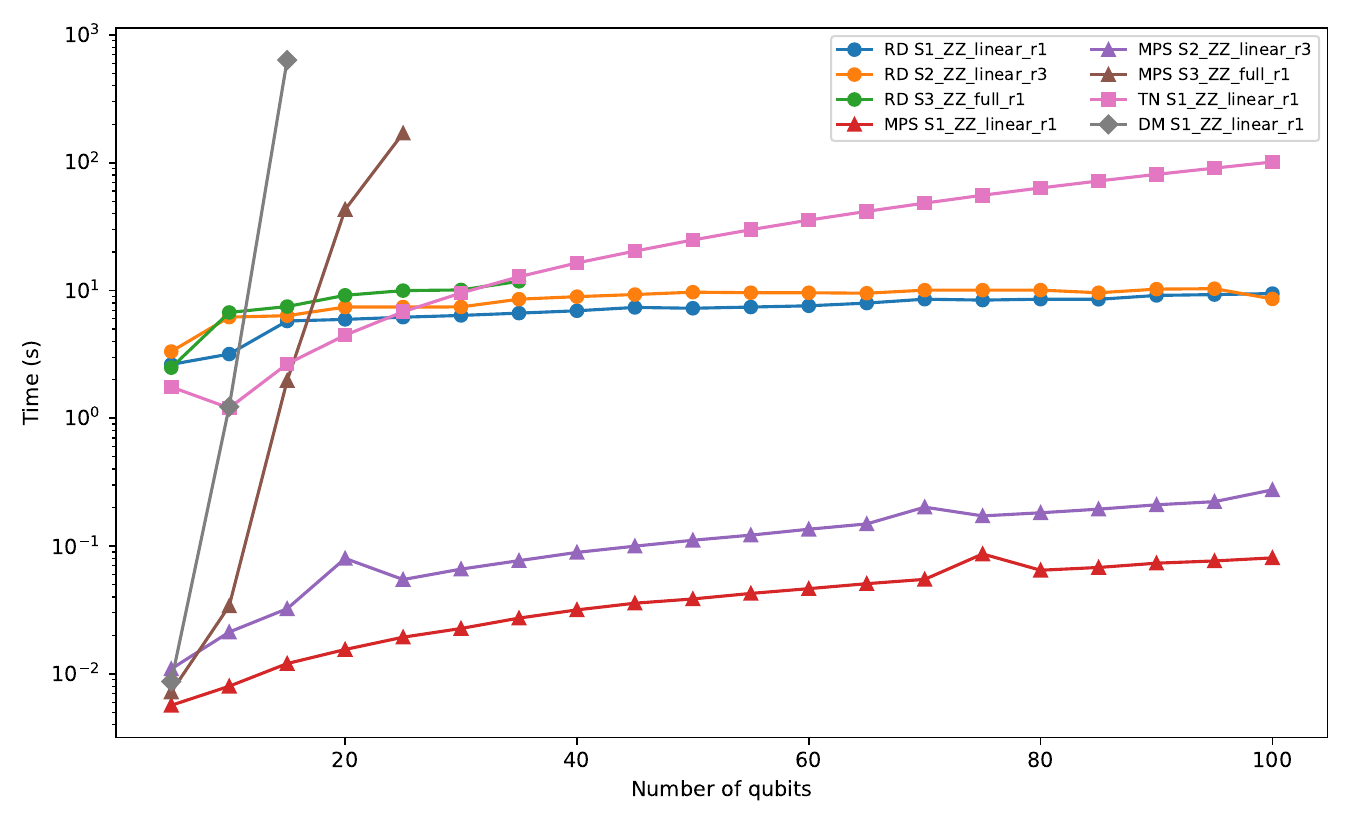}
\caption{Scalability of GSI runtime based on the number of qubits and the method used.}
\label{fig:scalability}
\end{figure}

This section analyzes how the computational cost of calculating GSI values grows with circuit size, to identify which execution environments remain practical as the number of qubits and gates increases. The execution environments considered in this study were introduced in Section~\ref{sec:foundationsScalability}. Figure~\ref{fig:scalability} summarizes their empirical scalability behavior.

The curves are grouped by circuit configuration, following the notation $S_k\_ZZ\_\{linear,full\}\_r\{1,3\}$. Here, $ZZ$ indicates that the entanglement block is based on $Z \otimes Z$ interactions, while \textit{linear} denotes a simple entanglement topology and \textit{full} refers to a fully connected entanglement topology. The suffixes $r1$ and $r3$ indicate the number of repetitions (layers) of the characteristic map block, that is, one and three repetitions, respectively, which increases the circuit depth and the total number of two-qubit interactions. For clarity, the three configurations analyzed in the scalability study are denoted as follows: S1 corresponds to $ZZ\_linear\_r1$, S2 corresponds to $ZZ\_linear\_r3$, and S3 corresponds to $ZZ\_full\_r1$.

First, the scalability results of the group of methods that run the simulations/emulations are discussed. If we examine the DM method line group, a rapid increase in memory consumption and computational cost as the number of qubits grows is observed (since it requires building and manipulating full-density matrices). For this reason, we evaluated two more scalable simulation strategies, MPS and TN. In both cases, the central idea is to avoid an explicit representation of the density matrix and, instead, work with factorizations that take advantage of the circuit structure, contracting only the tensors necessary to quantify the GSI metric. This substantially reduces the resources required compared to DM, and this difference is reflected in the fact that DM is not capable of exceeding 20 qubits. MPS and TN can be applied up to 100 qubits for the simplest configuration.

While MPS and TN clearly extend the simulation range, their scalability depends heavily on the circuit configuration. The linear entanglement pattern leads to more favorable growth than the full pattern, as the latter increases the effective connectivity and complexity of the contraction. Similarly, increasing the number of repetitions from r1 to r3 deepens the circuit and increases the number of two-qubit ZZ interactions, resulting in higher contraction costs and longer execution times for both simulators. This is consistent with the observed trends, as for sparse connectivity and shallow depth, the simulators scale smoothly, while denser entanglement and deeper circuits cause a more pronounced increase.

After analyzing the simulators, we now focus on the results from actual devices. Again, in Figure \ref{fig:scalability}, RD remains within a relatively narrow range across all explored qubit counts and configurations, indicating that the cost of running on the device does not degrade as aggressively as the cost of classical simulation when the system size is scaled up. Therefore, while MPS and TN are essential for driving experimentation to larger instances and studying how circuit structure affects simulation complexity, RD shows practical scalability, as it reflects runtime behavior in real hardware environments.

\section{Conclusions}\label{sec:conclusion}
This work presents GATE, a methodology for optimizing quantum circuits through the proposed GSI, a metric that quantifies the contribution of individual gates based on fidelity, entanglement, and sensitivity. By systematically removing low-contribution gates, GATE enables the construction of more efficient quantum feature maps while preserving, and often improving, predictive performance. Experimental results on multiple datasets, two representative QML models (PegasosQSVM and QNN), and different execution environments show that moderate circuit reduction consistently achieves a favorable balance between accuracy and runtime. In particular, the best-performing configurations are typically found at intermediate thresholds, rather than in the original circuits or in aggressively simplified ones. This confirms that reducing circuit complexity can mitigate noise accumulation and enhance effective model performance. Moreover, GATE provides an interpretable and systematic approach to circuit design, enabling gate-level analysis of quantum models and offering practical guidance for feature map construction. Its compatibility with both simulation and real hardware further supports its applicability in realistic quantum computing scenarios.

Despite these strengths, some limitations remain. The accuracy of the GSI depends on the quality of metric estimation, particularly in noisy hardware settings, and the method is primarily suited to quantum machine learning circuits where redundancy is present. In addition, the approach assumes that gate contributions can be assessed independently, which may not fully capture complex interactions in highly entangled circuits, where the impact of removing a gate may influence distant parts of the circuit.

Future work will focus on adaptive threshold selection, improved estimation techniques under noise, and integration with complementary optimization strategies. In particular, incorporating hardware-aware criteria into the threshold selection process could further improve robustness under noise conditions. Extending the GSI formulation to account for interactions between groups of gates, rather than evaluating them only individually, represents another promising direction. Additionally, exploring dynamic weighting schemes for the GSI components may allow the methodology to better adapt to different datasets and model requirements. Extending the methodology to more complex models and larger-scale quantum systems, as well as to multi-class settings and hybrid quantum-classical architectures, will further clarify its role in enabling scalable QML.

\section*{Supplementary Material}

This section provides the link to the GitHub repository, which includes the source code, datasets, images of the original and optimized circuits, detailed results tables, and instructions for reproducing the experiments: \href{https://github.com/Data-Science-Big-Data-Research-Lab/GATE_GSI}{https://github.com/Data-Science-Big-Data-Research-Lab/GATE\_GSI}

\section*{Acknowledgements}

The authors would like to thank the Spanish Ministry of Science and Innovation for the support within the projects PID2023-146037OB-C21, PID2023-146037OB-C22. We also acknowledge Pablo de Olavide University for funding the Q-Resilience project. Finally, we acknowledge the use of IBM Quantum Credits for this work. The views expressed are those of the authors and do not reflect the official policy or position of IBM or the IBM Quantum team.

\bibliographystyle{amsplain}
\bibliography{bibliography}

@book{Nielsen10,
  title={Quantum Computation and Quantum Information},
  publisher ={Cambridge University Press},
  author={Nielsen, Michael A. and Chuang, Isaac L.}, 
  pages = {1-702},
  year={2010},

}

@article{Preskill18,
  title={Quantum Computing in the NISQ era and beyond},
  author={Preskill, J.},
  journal={Quantum},
  volume={2},
  pages={79},
  year={2018}
}

@article{Barenco95,
  title={Elementary gates for quantum computation},
  author={Barenco, A. and Bennett, C. H. and Cleve, R. and others},
  journal={Physical Review A},
  volume={52},
  number={5},
  pages={3457-3467},
  year={1995}
}

@article{Devitt13,
  title={Quantum error correction for beginners},
  author={Devitt, S. J. and Munro, W. J. and Nemoto, K.},
  journal={Reports on Progress in Physics},
  volume={76},
  number={7},
  pages={076001},
  year={2013}
}

@article{Maslov17,
  title={Basic circuit compilation techniques for an ion-trap quantum machine},
  author={Maslov, D.},
  journal={New Journal of Physics},
  volume={19},
  number={2},
  pages={023035},
  year={2017}
}

@article{Kandala19,
  title={Error mitigation extends the computational reach of a noisy quantum processor},
  author={Kandala, A. and Temme, K. and C{\'o}rcoles, A. D. and Mezzacapo, A. and Chow, J. M. and Gambetta, J. M.},
  journal={Nature},
  volume={567},
  number={7749},
  pages={491-495},
  year={2019}
}

@article{Amy13,
  title={A meet-in-the-middle algorithm for fast synthesis of depth-optimal quantum circuits},
  author={Amy, M. and Maslov, D. and Mosca, M. and Roetteler, M.},
  journal={IEEE Transactions on Computer-Aided Design of Integrated Circuits and Systems},
  volume={32},
  number={6},
  pages={818-830},
  year={2013}
}

@article{Shende06,
  title={Synthesis of quantum-logic circuits},
  author={Shende, V. V. and Bullock, S. S. and Markov, I. L.},
  journal={IEEE Transactions on Computer-Aided Design of Integrated Circuits and Systems},
  volume={25},
  number={6},
  pages={1000-1010},
  year={2006}
}

@article{Temme17,
  title={Error mitigation for short-depth quantum circuits},
  author={Temme, K. and Bravyi, S. and Gambetta, J. M.},
  journal={Physical Review Letters},
  volume={119},
  number={18},
  pages={180509},
  year={2017}
}

@article{Li17,
  title={Efficient variational quantum simulator incorporating active error minimization},
  author={Li, Y. and Benjamin, S. C.},
  journal={Physical Review X},
  volume={7},
  number={2},
  pages={021050},
  year={2017}
}

@inproceedings{Murali19,
  title={Noise-adaptive compiler mappings for noisy intermediate-scale quantum computers},
  author={Murali, P. and Baker, J. M. and Javadi-Abhari, A. and Chong, F. T. and Martonosi, M.},
  booktitle={Proceedings of the Twenty-Fourth International Conference on Architectural Support for Programming Languages and Operating Systems},
  pages={1015-1029},
  year={2019}
}

@inproceedings{Siraichi18,
  title={Qubit allocation},
  author={Siraichi, M. Y. and Dos Santos, V. F. and Collange, C. and Pereira, F. M. Q.},
  booktitle={Proceedings of the International Symposium on Code Generation and Optimization},
  pages={113-125},
  year={2018}
}

@article{Mitarai18,
  title={Quantum circuit learning},
  author={Mitarai, K. and Negoro, M. and Kitagawa, M. and Fujii, K.},
  journal={Physical Review A},
  volume={98},
  number={3},
  pages={032309},
  year={2018}
}

@article{Bravy22,
  title={Hybrid quantum-classical algorithms for approximate graph coloring},
  author={Bravy, S. and Kliesch, A. and Koenig, R. and Tang, E.},
  journal={Quantum},
  volume={6},
  pages={678},
  year={2022}
}

@article{DeJong20,
  title={Embracing co-design for quantum computers and applications},
  author={De Jong, W. A. and {\it et al.}},
  journal={IEEE Transactions on Quantum Engineering},
  volume={1},
  pages={1-19},
  year={2020}
}

@article{Coles18,
  title={Quantum algorithm implementations for beginners},
  author={Abhijith, J. and Adedoyin, A. A. and Ambrosiano, J. J. and others},
  journal={ACM Transactions on Quantum Computing},
  volume={3},
  number={4},
  pages={1-92},
  year={2022}
}

@article{Schuld15,
  title={An introduction to quantum machine learning},
  author={Schuld, M. and Sinayskiy, I. and Petruccione, F.},
  journal={Contemporary Physics},
  volume={56},
  number={2},
  pages={172-185},
  year={2015}
}

@article{Maslov08,
  title={Comparison of the cost metrics for reversible and quantum logic synthesis},
  author={Maslov, D. and Miller, D. M.},
  journal={ET Computers \& Digital Techniques},
  volume={1},
  number={2},
  pages={98-104},
  year={2007}
}

@article{Nam18,
  title={Automated optimization of large quantum circuits with continuous parameters},
  author={Nam, Y. and Ross, N. J. and Su, Y. and Childs, A. M. and Maslov, D.},
  journal={npj Quantum Information},
  volume={4},
  pages={23},
  year={2018}
}

@article{Khatri19,
  title={Quantum-assisted quantum compiling},
  author={Khatri, S. and LaRose, R. and Poremba, A. and Cincio, L. and Sornborger, A. T. and Coles, P. J.},
  journal={Quantum},
  volume={3},
  pages={140},
  year={2019}
}

@article{Olson2017PMLB,
  author={Olson, R. S. and La Cava, W. and Orzechowski, P. and Urbanowicz, R. J. and Moore, J. H.},
  title={PMLB: a large benchmark suite for machine learning evaluation and comparison},
  journal={BioData Mining},
  year={2017},
  volume={10},
  pages={36}
}

@misc{dosad2024fitness,
  title={Fitness Club Dataset for ML Classification},
  author={Dosad, D.},
  year={2024},
  url={https://www.kaggle.com/datasets/ddosad/datacamps-data-science-associate-certification}
}

@misc{fedesoriano2021heart,
  title={Heart Failure Prediction},
  author={F. Soriano},
  year={2021},
  url={https://www.kaggle.com/datasets/fedesoriano/heart-failure-prediction}
}

@article{ANDERS10,
  title={Ancilla-driven quantum computation},
  author={J. Anders and D. K. L. Oi and E. Kashefi and D. E. Browne and E. Andersson},
  journal={Physical Review Letters},
  volume={82},
  pages={020301},
  year={2010}
}

@article{XU13,
  title={Gate sequence for continuous variable one-way quantum computation},
  author={X. Su and S. Hao and X. Deng and L. Ma and M. Wang and X. Jia and C. Xie and K. Peng},
  journal={Nature Communications},
  volume={82},
  pages={020301},
  year={2013}
}

@article{GRIFFITHS96,
  title={{Semiclassical Fourier Transform for Quantum Computation}},
  author={Griffiths, R. B. and Niu, C.},
  journal={Physical Review Letters},
  volume={76},
  number={17},
  pages={3228--3231},
  year={1996}
}

@article{SCHULD17,
  title={Implementing a distance-based classifier with a quantum interference circuit},
  author={Schuld, M. and Fingerhuth, M. and Petruccione, F.},
  journal={Europhysics Letters},
  volume={119},
  number={6},
  pages={60002},
  year={2017}
}

@article{VADALI24,
  title={{Quantum circuit fidelity estimation using machine learning}},
  author={Vadali, A. and Kshirsagar, R. and Shyamsundar, P. and Perdue, G. N.},
  journal={Quantum Machine Intelligence},
  volume={6},
  pages={1},
  year={2024}
}

@article{PERETZ24,
  title={{A parameterized quantum circuit for estimating distribution measures}},
  author={Peretz, O. and Koren, M.},
  journal={Quantum Machine Intelligence},
  volume={6},
  pages={22},
  year={2024}
}

@article{SIMON24,
  title={{On neural quantum support vector machines}},
  author={Simon, L. and Radons, M.},
  journal={Quantum Machine Intelligence},
  volume={6},
  pages={3},
  year={2024}
}

@article{RATH24,
  title={{Quantum data encoding: a comparative analysis of classical-to-quantum mapping techniques and their impact on machine learning accuracy}},
  author={Rath, M. and Date, H.},
  journal={EPJ Quantum Technology},
  volume={11},
  pages={72},
  year={2024}
}

@article{KOOKANI24,
  title={{XpookyNet: advancement in quantum system analysis through convolutional neural networks for detection of entanglement}},
  author={Kookani, A. and Mafi, Y. and Kazemikhah, P.n and Aghababa, H. and Fouladi, K. and Barati, M.},
  journal={Quantum Machine Intelligence},
  volume={6},
  pages={50},
  year={2024}
}

@article{BU24,
  title={{Complexity of Quantum Circuits via Sensitivity, Magic, and Coherence}},
  author={Bu, K. and Garcia, R. J. and Jaffe, A. and Koh, D. E. and Li, L.},
  journal={Communications in Mathematical Physics},
  volume={405},
  pages={161},
  year={2024}
}

@article{RODRIGUEZ26,
  title={{A Survey of Quantum Machine Learning: Foundations, Algorithms, Frameworks, Data and Applications}},
  author={F. Rodríguez-Díaz and D. Gutiérrez-Avilés and A. Troncoso and F. Martínez-Álvarez},
  journal={ACM Computing Surveys},
  volume={58},
  issue={4},
  pages={1-35},
  year={2026}
}

@article{Sun25,
  title={{MLQM: Machine learning approach for accelerating optimal qubit mapping}},
  author={Sun, W. and Li, X. and Yu L. and Wang, Z. and Chen, G. and Yang, G.},
  journal={Future Generation Computer Systems},
  volume={173},
  pages={107906},
  year={2025}
}

@article{Rattacaso25,
  title={{Quantum circuit compilation with quantum computers}},
  author={Rattacaso, D. and Jaschke, D. and Ballarin, M. and Siloi, I. and Montangero, S.},
  journal={Physical Review Research},
  volume={7},
  pages={033268},
  year={2025}
}

@article{Huang25,
  title={{Qubit mapping: the adaptive divide-and-conquer approach}},
  author={Huang, Y. and Zhou, X. and Meng, F. and Li, S.},
  journal={Quantum Information Processing},
  volume={24},
  pages={202},
  year={2025}
}

@article{Valois26,
  title={{Efficient and scalable branch-and-bound algorithm for exact qubit allocation}},
  author={Valois, J. P. and Helbecque, G. and Melab, N},
  journal={Future Generation Computer Systems},
  volume={179},
  pages={108342},
  year={2025}
}

@article{Yale25,
  title={{Noise-aware circuit compilations for a continuously parameterized two-qubit gateset}},
  author={Yale, C. G. and Rines, R. and Omole, V. and Thotakura, B. and Burch, A. D. and Chow, M. N. H. and Ivory, M. and Lobser, D. and McFarland, B. K. and Revelle, M. C. and Clark, S. M. and Gokhale, P.},
  journal={Physical Review Applied},
  volume={24},
  pages={024057},
  year={2025}
}

@inproceedings{Ramadhani23,
    author  = {Ramadhani, M. H. and Abdurohman, M. and Nuha, H. H.},
    booktitle={Proceedings of the International Conference on Data Science and Its Applications},
    title   = {{Dynamical Decoupling (DD) to Improve Fidelity in Quantum Computing}},
    year    = {2023},
    volume  = {},
    pages   = {495-499}
}

@article{Hirai24,
  title={{Practical application of quantum neural network to materials informatics}},
  author={Hirai, H.},
  journal={Scientific Reports},
  volume={14},
  pages={8583},
  year={2024}
}

\end{document}